\documentclass{article}

\usepackage{microtype}
\usepackage{graphicx}
\usepackage{booktabs} 
\usepackage{multirow}
\usepackage{makecell}

\usepackage{adjustbox}
\usepackage{colortbl} 
\usepackage{xcolor}
\usepackage{listings}
\usepackage{fontawesome}
\usepackage{pifont}

\usepackage{hyperref}
\usepackage{url}

\usepackage[accepted]{icml2024}

\usepackage{amsmath}
\usepackage{amssymb}
\usepackage{mathtools}
\usepackage{amsthm}

\usepackage{placeins}

\usepackage[capitalize,noabbrev]{cleveref}
\Crefname{section}{\mbox{\S\hspace*{-0.25ex}}}{\mbox{\S\hspace*{-0.25ex}}}
\Crefname{appendix}{\S$\!$}{\S$\!$}

\usepackage{caption}
\usepackage{enumitem} 
\setlist[itemize]{noitemsep,topsep=0ex}    
\setlist{leftmargin=3mm}

\usepackage{array}
\newcolumntype{P}[1]{>{\raggedright\arraybackslash}p{#1}}
\newcolumntype{M}{>{\ttfamily}l} 

\usepackage[light,scaled=0.825]{roboto-mono}

\usepackage{tikz}
\newcommand{\circledigit}[1]{%
\tikz[baseline=(char.base)]{
    \node[shape=circle,draw,inner sep=1pt, fill=red, text=white] (char) {\footnotesize\textsf{\textbf{#1}}};
}}

\theoremstyle{plain}

\theoremstyle{definition}

\theoremstyle{remark}

\usepackage{subcaption}

\usepackage[subtle]{savetrees}

\newcommand{\paragraphtight}[1]{\textbf{#1~}}

\newcommand{\wataskcount}{33}
\newcommand{\wainstancecount}{{19,912}}

\icmltitlerunning{WorkArena: Web Agents for Common Knowledge Work Tasks}

\begin{document}

\twocolumn[
\icmltitle{WorkArena: How Capable are Web Agents at\texorpdfstring{\\}{ }Solving Common Knowledge Work Tasks?}

\icmlsetsymbol{equal}{*}
\icmlsetsymbol{core}{$\dagger$}

\begin{icmlauthorlist}
\icmlauthor{Alexandre Drouin}{equal,snow,mila}
\icmlauthor{Maxime Gasse}{equal,snow,mila,poly}
\icmlauthor{Massimo Caccia}{core,snow}
\icmlauthor{Issam H. Laradji}{snow}\\
\icmlauthor{Manuel Del Verme}{mila,mcgill}
\icmlauthor{Tom Marty}{snow,mila,udem}
\icmlauthor{Léo Boisvert}{snow,poly}
\icmlauthor{Megh Thakkar}{snow,mila,udem}
\icmlauthor{Quentin Cappart}{mila,poly}\\
\icmlauthor{David Vazquez}{snow}
\icmlauthor{Nicolas Chapados}{snow,mila,poly}
\icmlauthor{Alexandre Lacoste}{core,snow}
\end{icmlauthorlist}

\icmlaffiliation{snow}{ServiceNow Research}
\icmlaffiliation{mila}{Mila -- Quebec AI Research Institute}
\icmlaffiliation{mcgill}{McGill University}
\icmlaffiliation{udem}{Université de Montréal}
\icmlaffiliation{poly}{Polytechnique Montréal}

\icmlcorrespondingauthor{Alexandre Drouin}{alexandre.drouin@servicenow.com}
\icmlcorrespondingauthor{Maxime Gasse}{maxime.gasse@servicenow.com}

\icmlkeywords{Machine Learning, ICML}

\vskip 0.3in
]



\printAffiliationsAndNotice{\icmlEqualContribution \textsuperscript{$\dagger$}Core contributor } 

\begin{abstract}
We study the use of large language model-based agents for interacting with software via web browsers. Unlike prior work, we focus on measuring the agents' ability to perform tasks that span the typical daily work of knowledge workers utilizing enterprise software systems. To this end, we propose WorkArena, a remote-hosted benchmark of 33 tasks based on the widely-used ServiceNow platform.  We also introduce BrowserGym, an environment for the design and evaluation of such agents, offering a rich set of actions as well as multimodal observations. Our empirical evaluation reveals that while current agents show promise on WorkArena, there remains a considerable gap towards achieving full task automation. Notably, our analysis uncovers a significant performance disparity between open and closed-source LLMs, highlighting a critical area for future exploration and development in the field.
\end{abstract}

\begin{figure*}[t]
    \centering

    \begin{subfigure}{0.57\textwidth}
      \centering
      \includegraphics[width=.97\linewidth]{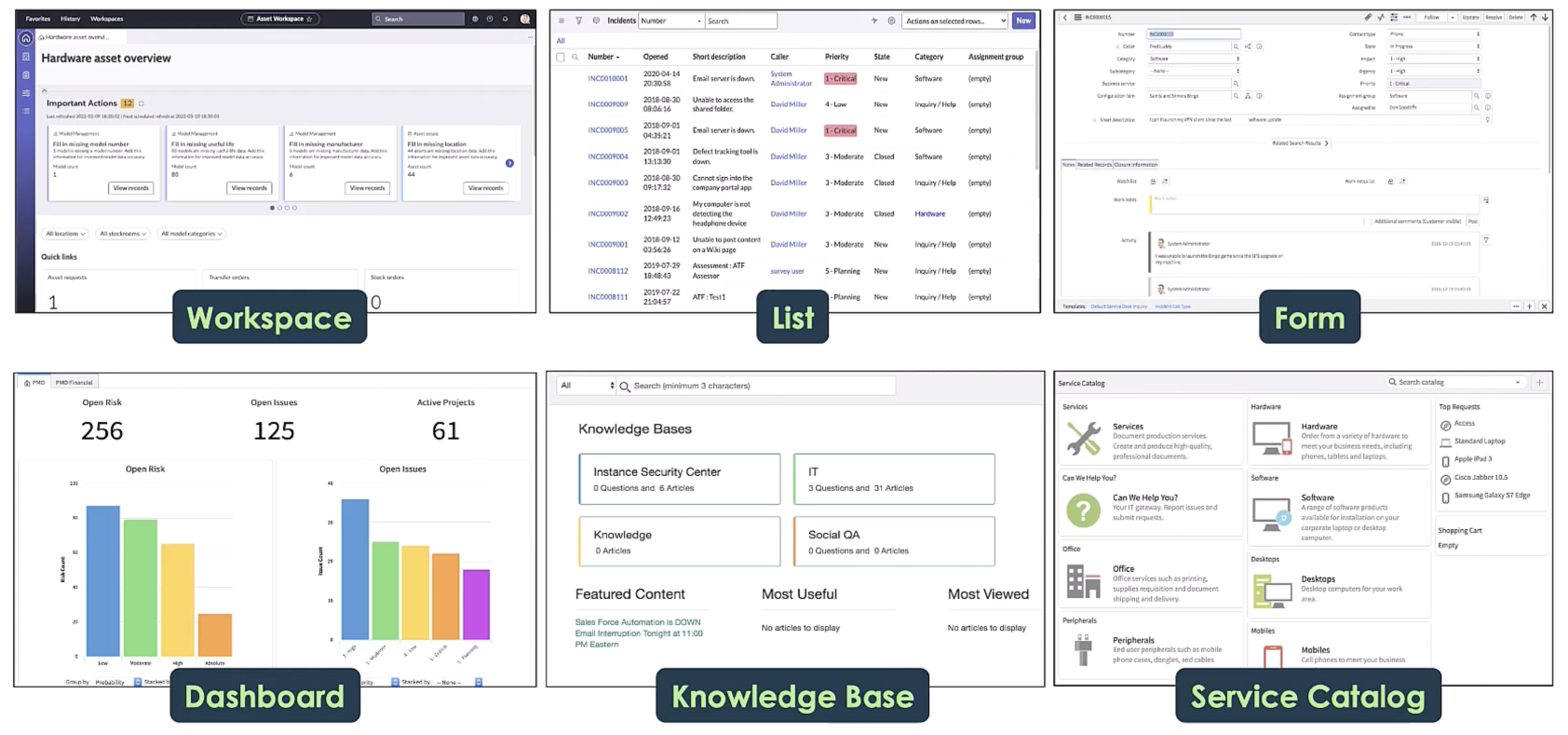}  
      \caption{WorkArena}
      \label{fig:workarena}
    \end{subfigure}%
    \hfill%
    \begin{subfigure}{0.43\textwidth}
      \centering
      \includegraphics[width=.97\linewidth]{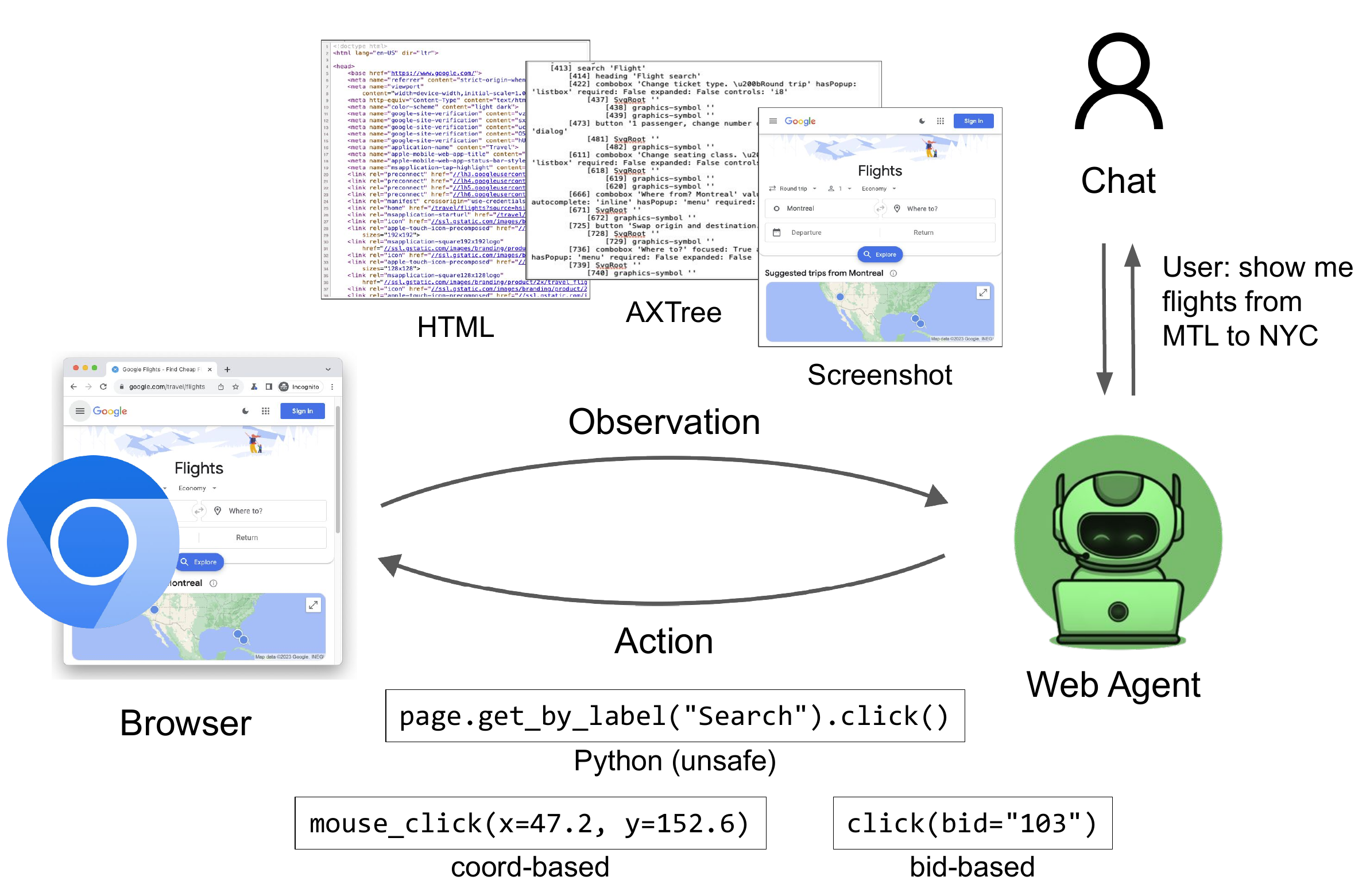}  
      \caption{BrowserGym}
      \label{fig:browsergym}
    \end{subfigure}
    
    \caption{Overview of contributions: (a) \textbf{WorkArena} is a benchmark of \wataskcount~web tasks and \wainstancecount~unique instances that cover common ways of interacting with the ServiceNow Platform, a widely-used enterprise software platform. (b) \textbf{BrowserGym} is a Python environment for designing and evaluating web agents, which includes a rich set of actions and multimodal observations (shown here the HTML contents of the page, its accessibility tree, and the raw pixels after browser rendering).}
    \label{fig:workarena_browsergym}
\end{figure*}

\section{Introduction}

Graphical User Interfaces (UIs) are the predominant medium through which people interact with software, serving as a crucial gateway to the digital world. 
While they have evolved to become more intuitive, featuring generic and universal components like forms, lists, and buttons, UIs can still make complex or repetitive tasks burdensome for users. While being more and more intuitive, those complex UIs also became inadvertedly more and more discriminative for visually-impaired users.
An ideal user experience would involve automated assistants that can streamline these tasks ensuring accessibility for everyone.
While Application Programming Interfaces (APIs) have facilitated programmatic interactions with software, the resulting automated assistants often lack transparency, are difficult for users to inspect, and are not universally available.
In contrast, assistants that directly manipulate UIs (UI assistants) offer greater transparency and are more amenable to human oversight.
Most notably, because the user can give and take back control over the UI at any point, UI assistants can provide varying levels of automation ranging from partial assistance (such as finding a menu or filling a form) to complete task execution (like placing an order), akin to the six levels of automation in autonomous driving~\citep{sae/levels-automation}.

Recent advancements in the fields of large language and vision models have seen the rapid development of UI assistants, particularly \emph{web agents} acting through a browser~\citep{furuta2023multimodal,kim2023rciagent,gur2023webagent}. The range of web tasks explored in the literature varies from simple UI commands such as selecting specific menu elements on toy web pages~\citep{LiuL18MiniWoB, Shi2017MiniWoB}, to more complex requests such as ``Checkout merge requests assigned to me'', on real-world websites like Reddit and GitLab~\citep{zhou2023webarena}. Yet, one area in which web agents can be particularly impactful and remain unexplored is enterprise software. In the workplace, where repetitive tasks are common, enterprise software often prioritizes functionality over user experience, leading to inefficiencies and long learning curves for workers.
Our work addresses this gap, and investigates the potential of web agents in enterprise settings to improve accessibility, user experience, and worker productivity.

To this end, we introduce \emph{WorkArena}, a benchmark developed on the widely-used ServiceNow platform~\citep{servicenow/vancouver}. ServiceNow is a comprehensive cloud-based platform that offers solutions for automating and managing digital workflows across various enterprise functions, including IT service management, human resources, customer service, and security operations.
In 2023 their customer base counted over 7,000 companies worldwide, including 85\% of the Fortune 500 companies~\citep{fortune500_2023}. 
Within these firms alone, the ServiceNow platform potentially impacts over 12 million individuals, not including broader public interactions,
such as the 500,000 daily users of Disney+'s customer help center~\citep{maas2020knowledge}.
ServiceNow's extensive reach makes it an ideal real-world environment for evaluating the potential impact of UI assistants in the workplace.

Our contributions are as follows:
\begin{itemize}
    \item \textbf{WorkArena:} A realistic benchmark of enterprise-related tasks for web agents comprising \wainstancecount\ unique task instances (\cref{sec:workarena}, \cref{fig:workarena_browsergym}a);
    \item \textbf{BrowserGym:} A new framework for the development and evaluation of web agents, compatible with previous benchmarks like WebArena~\citep{zhou2023webarena}, MiniWoB~\citep{LiuL18MiniWoB, Shi2017MiniWoB} and WebShop~\citep{yao2022webshop}, that offers a richer set of multimodal observations (e.g., screenshot, accessibility tree, screen coordinates), a broader set of actions (e.g., Python code and high-level primitives), and supports chat-based interactions (\cref{sec:browsergym}, \cref{fig:workarena_browsergym}b); Surprisingly, these features contribute to bringing our GPT-4 agent at the top of the leaderboard on WebArena, with a score of 25.4\%, contrasting with the score of the original paper, 14.4\%.
    \item \textbf{Empirical study:} We report a collection of experiments to assess the ability of state-of-the-art large language model (LLM)-based agents to solve WorkArena, as well as an analysis of the impact of the different BrowserGym features on WorkArena and MiniWoB (\cref{sec:results}).
\end{itemize}

\section{Related Works}

\paragraphtight{Benchmarks for web agents:}
Early benchmarks for web agents were based on synthetic web environments where agents were tasked with performing low-level keyboard and mouse actions~\citep{shi2017world}. Notable examples are MiniWoB~\citep{Shi2017MiniWoB, LiuL18MiniWoB}, which offer a collection of 125 toy web tasks ranging from clicking a specific button to using a basic text editor, and WebShop~\citep{yao2022webshop}, a simulated e-commerce website with shopping tasks that require searching and browsing a catalog of items. More recently, \citet{zhou2023webarena} introduced WebArena, a collection of 190 tasks based on realistic websites that emulate real-world domains such as e-commerce, social forums, collaborative software development, and content management. WebArena is a notoriously challenging benchmark, with a success rate of 14\% for a state-of-the-art web agent based on GPT-4, and 78\% for human agents.
\citet{deng2023mind2web} proposed Min2Web, a large-scale dataset of 2,000 web interactions from 137 websites curated by human annotators.
Similarly, \citet{lu2024weblinx} propose WebLINX, a curated dataset of web interactions composed of 2337 expert demonstrations from 155 different real-world websites. In WebLINX, each task is composed of a turn-based chat dialogue averaging 43 interactions per task. 
Recently, \citet{he2024webvoyager} propose 300 information-retrieval tasks, from 15 real-world consumer websites (e.g., Amazon, Coursera, Booking), which are used to evaluate WebVoyager, a vision-based web agent.

Worth mentioning is the related body of work on mobile UI agent benchmarks, which includes the early PixelHelp dataset \citep{seq2act} and more recent Android in the Wild \citep{aitw23} and Macro Mining \citep{macromining}.

Our proposed benchmark, WorkArena, is designed to complement existing work by specifically focusing on real-world enterprise software applications. It includes a wide range of tasks that collectively encompass several end-to-end workflows typically performed by knowledge workers.
Additionally, it poses a series of technical challenges, such as pages with very large document object models (DOMs), non-standard HTML, and complex UI elements, which we outline in \cref{sec:workarena_challenges}. This benchmark integrates into BrowserGym, a new environment that we propose for the evaluation of web agents, which aggregates all features proposed in previous work, such as multimodal observations and code-based actions while being the first to support chat-based agent-user interactions (\cref{sec:browsergym-features}).

\paragraphtight{LLM-based Agents:} The scope of our experimental contributions is limited to web agents that rely on language models for reasoning.
Recent studies include the seminal work of \citet{nakano2021webgpt} that introduces WebGPT, an agent capable of browsing the web and answering questions via information retrieval.
Other works have also explored web agents that receive HTML as input and produce a series of high-level actions such as \emph{click}, \emph{type}, \emph{select}~\citep{deng2023mind2web, liu2023agentbench,liu2023bolaa,yao2023react}.

Other works have shown that using only textual information as input is sometimes limiting and have considered multimodal observations that combine visual information (screenshots of a page) with text~\citep{Humphreys22rl4MiniWoB, he2024webvoyager}.
Instead of directly interacting with a website, recent works have proposed methods that can act on websites using Python-generated code from task-specific instructions \citep{gur2023real,gur2023webagent}.
Our proposed environment, BrowserGym is flexible in that it supports all observations and actions spaces utilized in prior research.

\section{WorkArena -- An Enterprise Benchmark}\label{sec:workarena}

WorkArena consists of a suite of \wataskcount~tasks and \wainstancecount\ unique instances that cover core interactions with the ServiceNow platform, such as filtering lists, filling forms, searching knowledge bases, utilizing service catalogs, reading dashboards, and navigating the workspace via menus (see \cref{fig:workarena}). Collectively, these tasks are representative of a wide array of common operations that employees, like IT, administrative, and white-collar staff, perform on a daily basis.

As a guiding example, consider an IT support agent tasked with onboarding new hires. Each day, this agent logs into the ServiceNow platform. On the landing page, they see a \textbf{dashboard} showing key statistics of the work assigned to them. Using the workspace's \textbf{menu}, they navigate to a list of requests to be fulfilled. They then \textbf{filter the list} to extract all requests assigned to them and \textbf{sort} them by priority. Finally, they process each request by \textbf{filling out forms} to create new user profiles and using the \textbf{service catalog} to order laptops for them.
As we will see, all of the interactions listed above are included in WorkArena, and this is only one of the many realistic user trajectories that the benchmark covers.

\begin{figure}
    \centering
    \includegraphics[width=0.77\linewidth]{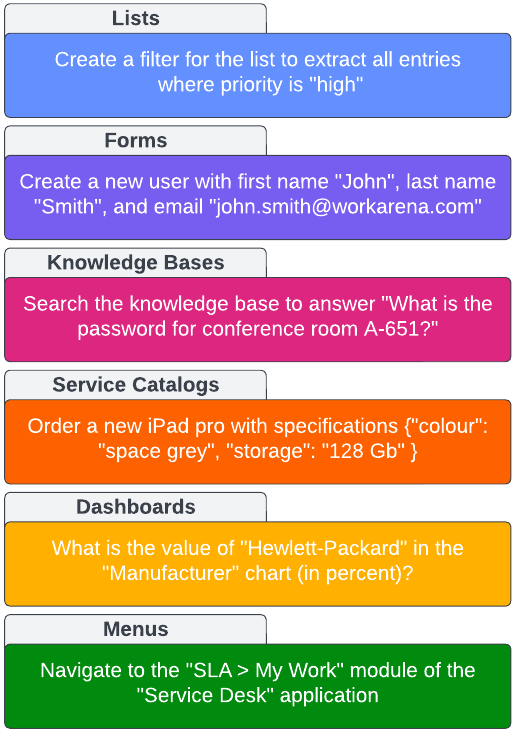}
    \caption{Example goal for each kind of WorkArena task.}
    \label{fig:goal-examples}
\end{figure}

\subsection{WorkArena Tasks}

\begin{figure*}
    \centering
    \includegraphics[width=0.97\linewidth]{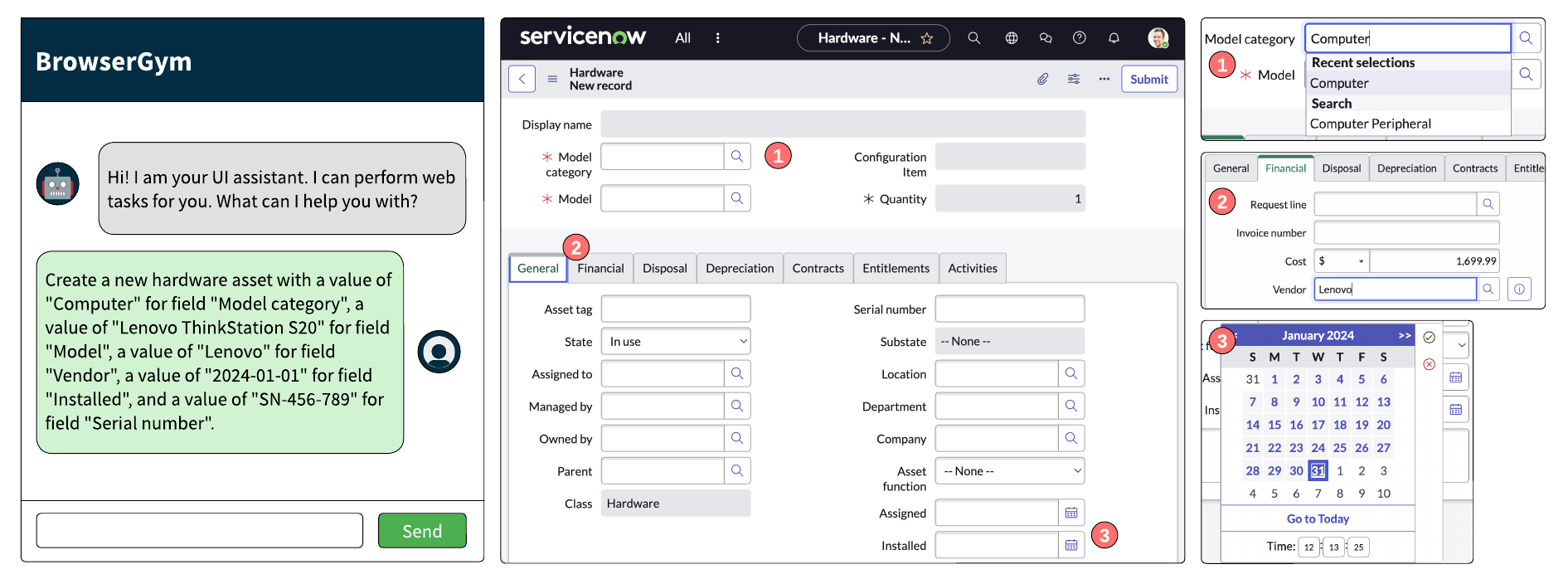}
    \caption{Example form task -- The goal is given to the agent in natural language via the chat interface. As can be seen, the goal is designed to be very explicit, leaving no ambiguity on the task to perform. As for the UI, it is complex, composed of many fields, some of which are dynamic, such as auto-completion-based text boxes~\circledigit{1}, some are hidden behind tabs~\circledigit{2}, and others require complex interactions, such as date pickers~\circledigit{3}. Other such examples are available in \cref{app:wa-task-ui}.}
    \label{fig:wa-form-trajectory}
\end{figure*}

In WorkArena, each task is coupled with a natural language goal that provides instructions to the agent. Each goal is automatically generated from a human-designed template filled with pre-defined values (menu name, field value, item specifications etc.), and explicitely provides all the information required to solve the task (examples in \cref{fig:goal-examples}). Other key components of tasks include: i) \emph{validation functions} and ii) \emph{oracle functions}, a unique feature that distinguishes WorkArena from previous work. The validation functions offer real-time feedback to agents, identifying errors ranging from minor (e.g., unfilled mandatory fields) to critical (such as pushing invalid data to the database).
The oracle functions are hand-crafted solutions that automatically complete the tasks using Playwright browser automation~\citep{Playwright}.
They serve three purposes: (i)~they ensure the feasibility of each task, (ii)~they act as ground truth for agents that have learning capabilities, and (iii)~they help maintain the benchmark's longevity by making it easier to identify and adjust tasks impacted by future updates to the ServiceNow platform.
Below, we outline the main categories of tasks included in the benchmark.

\paragraphtight{Lists:} We consider 12 list-based tasks, which can be grouped into two categories: filtering and sorting. The former consists of using the UI to construct a complex filter with 1 to 5 conditions. The latter consists of using the UI to sort the list based on up to 3 columns. In both cases, the interaction with the UI is non-trivial and requires opening a hidden menu, adding the right number of conditions, and filling them out accordingly. There are 6 tasks of each type, each corresponding to a different data table (e.g., users, incidents). In both cases, client-side validation is used to verify that the resulting list satisfies the expected conditions. Together, these tasks yield 6,900 instances.

\paragraphtight{Forms:} We consider 5 form-based tasks, which each consist of creating a new entry in a given data table. These tasks vary in complexity based on the number of fields that must be filled (from 1 to 26) and on the intricate properties of each form's UI. For example, some forms require navigating through a set of tabs to expose hidden fields. Others use dynamic auto-completions fields, which require careful handling (see \cref{fig:wa-form-trajectory} for an example). In all cases, validation proceeds by querying the database to retrieve entries created by the agent and verifying that their values are as expected. Together, these tasks yield 5,000 instances.

\paragraphtight{Knowledge bases:} The benchmark includes an information retrieval task that consists of searching the platform's knowledge base to answer a question. Concretely, this requires searching with appropriate keywords and browsing the resulting articles to find specific information. These tasks are constructed by starting from a list of facts, generating articles containing each fact with GPT-4~\citep{OpenAI2023GPT4TR}, and generating a series of questions that unambiguously ask for this fact. Then, validation proceeds by verifying if the answer returned by the agent is within a set of acceptable answers. For example, if the question is ``What is the level of customer satisfaction?'' and the answer is ``8.5/10'', alternative answers such as ``85\%'' or ``8.5 out of 10'' would be accepted. In total, this task yields 1,000 instances. Details on article generation and validation are given in \cref{app:wa-kb-task-details}.

\paragraphtight{Service catalogs:} The benchmark includes 9 tasks that require navigating a catalog of products and ordering items with given specifications. Such tasks vary in complexity based on the number of item configuration options. In all cases, validation is done by querying the database to verify that the order request created by the agent includes the right items in the right amounts, with the expected specifications. Together, these tasks yield 3,550 instances.

\paragraphtight{Dashboards:} The benchmark includes 4 retrieval tasks that require extracting numerical data from charts and (optionally) performing simple reasoning over them to answer a question. The complexity of such tasks depends on i) whether there are one or many charts on the page and ii) the reasoning to be done (e.g., extract the minimum).
In all cases, validation consists of verifying if the agent's response contains key numbers and labels referring to chart elements.
Together, these tasks yield 1,862 instances.

\paragraphtight{Menus:} We consider 2 menu-based tasks: i) navigating via the ``All'' menu and ii) impersonating users. The first consists of using the platform's main menu to navigate to a given application. In this case, validation simply verifies that the agent has arrived at the expected location. The second consists of \emph{impersonating} a user, a task commonly performed by IT support agents, where the agent logs into the platform as a given user to diagnose an issue. In this case, validation verifies that the expected user is logged in. Together, these tasks yield 1,600 instances.

A detailed list of all the tasks implemented in WorkArena is available in \cref{app:wa-tasks} \cref{tab:workarena_all_tasks}.

\subsection{Challenges: the World \emph{Wild} Web of Work}
\label{sec:workarena_challenges}

The ServiceNow platform poses a specific set of challenges for UI assistants, which we believe make WorkArena a complementary and meaningful benchmark for the community.

\textbf{Non-standard, dynamic UIs:} First, the web pages are heavily dynamic, and exhibit complex UI elements and ways of interacting with them. For example, the form to create an incident contains specific rules that can make some fields required or hidden, depending on the value of other fields (e.g., setting an incident's status to ``Resolved'' requires filling its ``Resolution notes''). Or, on some pages, the right-click can be overloaded to display a dynamic menu in certain areas. While these UI behaviors are not necessarily standard or best practices in web development, they are fairly common and representative of real-world enterprise software, which is not always designed with user accessibility in mind.

\textbf{Non-standard, exotic HTML:} Second, the ServiceNow platform relies on a complex combination of web technologies to implement its web pages, with nested iFrames, shadow DOMs, and proprietary Javascript APIs and HTML tags that do not necessarily adhere to existing web standards.\footnote{\url{https://www.w3.org/standards/}} This specificity would require strong out-of-distribution generalization for a UI assistant to successfully solve a task.

\textbf{Large HTML:} Third, the Document Object Model (DOM) of rendered web pages in the ServiceNow platform can be prohibitively large even for state-of-the-art language models, with a flat HTML text size that ranges between 40k and 500k tokens, even after a basic cleaning (removing scripts, styles and empty elements). Thus, even a conceptually simple task, such as finding the next element to click on the current page requires long context understanding, which is an active area of research in language models.

\subsection{Availability}

WorkArena is open-source and designed to be easily extended by the scientific community.\footnote{\url{https://github.com/ServiceNow/WorkArena}}
Tasks are executed on Personal Developer Instances, which are real cloud-based instances of the ServiceNow product, accessible for free through their \href{https://developer.servicenow.com}{developer program}.
It is important to note that although our benchmark is built on top of ServiceNow’s platform, it operates independently of any proprietary code.

\section{BrowserGym}\label{sec:browsergym}

Along with the WorkArena benchmark, we introduce BrowserGym, a generic browser environment that facilitates the design of new benchmarks, and provides a solid platform for the evaluation of multi-modal web agents.\footnote{\url{https://github.com/ServiceNow/BrowserGym}} BrowserGym (\cref{fig:browsergym}) is implemented as an OpenAI Gym environment~\citep{software/gym} and follows a Partially-Observable Markov Decision Process (POMDP) paradigm.\footnote{BrowserGym extracts at each time step an observation that is based only on the current view of the page. This emphasizes the need for web agents to implement some kind of memory mechanism, which BrowserGym does not provide.} It relies on Chromium and uses the Chrome DevTools Protocol (CDP)~\citep{software/chromedevtools} and the Playwright library~\citep{Playwright} to interact with the web browser.

\subsection{Capabilities}\label{sec:browsergym-features}

BrowserGym implements the following capabilities.

\textbf{Chat-based user interaction:} One of the interaction modalities is a chat interface where the user and the web agent can exchange messages. In WorkArena, the goal of each task is provided as the initial user message, to which the agent can reply at any time. This allows for information retrieval tasks where a specific answer is expected from the agent, but also sequential tasks where user instructions change over time and are delivered sequentially in a way that mimics real-world use cases.

\textbf{Augmented attributes:} For every element on the current page BrowserGym provides a unique identifier \verb|(bid)|, its bounding box coordinates \verb|(left,top,right,bottom)|, and flags indicating whether it is \verb|visible| and \verb|clickable|. These attributes both provide a crude summary of the visual rendering of the UI and allow for unambiguous interaction with individual elements using their identifiers.

\textbf{Rich observation space:} at each time step, the observation space contains the content of the chat (list of messages), the currently open pages (list of URLs), the error message from the last action if any (stack trace), and a multi-modal view of the active web page: its HTML DOM snapshot (structured object), its accessibility tree or AXTree (structured object) as originally proposed by~\citet{zhou2023webarena}, and a viewport screenshot (image). Both DOM and AXTree are obtained through Chrome's CDP, and are enriched with augmented attributes \verb|(bid, coords, visible, clickable)|. These structured objects can typically be rendered as text for processing by a language model and can be combined with the screenshot for a vision-language model.

\textbf{Rich action space:} the action space is customizable and includes Python code, which can be restricted to specific set of pre-defined high-level primitives, such as \verb|bid|-based actions (\verb|click(bid)|, \verb|type(bid,text)|, \dots), and \verb|coord|-based actions (\texttt{mouse\_click(x,y)}, \texttt{keyboard\_type(text)}, \dots). Alternatively, the action space can allow for the execution of arbitrary Python code, including the entire Playwright API which gives the web agent maximum flexibility in interacting with the browser.
For the complete list of high-level primitives available in Browsergym, refer to \cref{app:browsergym}, \cref{tab:browsergym_action_space}.

\paragraphtight{Multi-page navigation:} BrowserGym natively supports web tasks that require multiple open pages (tabs, popups) and is also robust to web pages that employ nested encapsulation techniques such as iFrames and shadow DOMs. This robustness is essential to handle the heterogeneity of real-world websites and is missing in existing web environments.

\subsection{An Ideal Experimental Framework}

\paragraphtight{Flexible agent design:} BrowserGym offers an extensive list of features but does not impose any restriction on how web agents should be implemented. The agent is responsible for using the provided observations or not (HTML, AXTree, screenshot, error message), deciding how to handle the history (past observations and actions), or deciding which action space it should be using (python, bid, coord, coord+bid). As such, with BrowserGym, researchers can easily experiment with new ideas and evaluate and compare a wide variety of web agents on the same set of tasks, such as text-only agents, vision-augmented agents, memory-augmented agents, and so on.

\paragraphtight{Minimal task design:} BrowserGym reduces the burden of creating new benchmarks to a minimum. Implementing a new task in BrowserGym boils down to implementing four functions: \texttt{\footnotesize setup()}, \texttt{\footnotesize teardown()}, \texttt{\footnotesize validate()} and \texttt{\footnotesize cheat()} (an optional oracle function). The \texttt{\footnotesize setup()} function is responsible for initializing anything the task needs beforehand, such as creating database entries, navigating to the starting URL, authenticating, etc. Likewise, \texttt{\footnotesize teardown()} is responsible for cleaning up any resource that might have been created during the task's execution. \texttt{\footnotesize validate()} is responsible for checking if the task's goal was fulfilled, which can involve operations such as querying a database, validating the URL and the content of the current page, or looking at the messages sent by the agent in the chat. The method returns a reward, an optional user message for the chat, and a \texttt{\footnotesize done} flag indicating the end of the task. Finally, each task can optionally implement an oracle function, \texttt{\footnotesize cheat()}, that automatically solves the task using a hard-coded Playwright solution. This can notably be used to assess the feasibility of tasks.

\paragraphtight{Extensibility:}
\href{https://github.com/ServiceNow/BrowserGym}{BrowserGym}
is easily extensible to additional benchmarks. We currently support MiniWoB \cref{app:miniwob}), WebArena \citep{zhou2023webarena} and WorkArena. We encourage the community to add new benchmarks or new agents to this platform. We also offer support to reduce the friction of adding new benchmarks to the platform. Please reach out on \href{https://github.com/ServiceNow/BrowserGym/discussions}{GitHub} for assistance.

\section{Experiments}
\label{sec:results}

We present a series of empirical experiments to assess the performance of state-of-the-art, general-purpose LLMs at solving work-related web tasks, using WorkArena and BrowserGym. The aim of these experiments is two-fold. First, we situate the level of difficulty of WorkArena by comparing it across baselines and benchmarks. Second, we propose an ablation study to quantify the impact of the different features offered in BrowserGym. All experiments are run with BrowserGym v0.3.5 and WorkArena v0.3.0.\footnote{Code: \url{https://github.com/ServiceNow/AgentLab}.}

\newcommand{\yes}{{\color{green}\ding{51}}}
\newcommand{\no}{{\color{red}\ding{55}}}

\begin{table}[ht]
\caption{Best agent configuration for each LLM, after a random search on MiniWoB and WorkArena.}
\noindent\resizebox{\columnwidth}{!}{
\begin{tabular}{l l c c c}
\toprule
\textbf{Type} & \textbf{Flag} & \textbf{GPT-4o} & \textbf{GPT-3.5} & \textbf{Llama3} \\
\midrule
\multirow{4}{*}{Agent}
& \texttt{use\_thinking} & \yes & \yes & \yes \\
& \texttt{use\_action\_history} & \yes & \yes & \yes \\
& \texttt{use\_error\_history} & \no & \no & \no \\
& \texttt{use\_think\_history} & \no & \no & \yes \\
\midrule
\multirow{6}{*}{Obs. Space}
& \texttt{use\_focused\_element} & \no & \no & \yes \\
& \texttt{use\_last\_error} & \yes & \yes & \no \\
& \texttt{coords} & \no & \no & \no \\
& \texttt{extract\_visible\_tag} & \yes & \yes & \yes \\
& \texttt{extract\_clickable\_tag} & \yes & \no & \no \\
& \texttt{only\_visible\_elements} & \no & \no & \no \\
\midrule
\multirow{4}{*}{Act. Space}
& \texttt{multi\_actions} & \no & \no & \no \\
& \texttt{action\_set} & \texttt{bid} & \texttt{bid} & \texttt{bid} \\
& \texttt{individual\_examples} & \yes & \yes & \yes \\
& \texttt{long\_description} & \yes & \no & \no \\
\bottomrule
\end{tabular}
}
\label{tab:best-agent-configs}
\end{table}

\subsection{Agent Design}\label{sec:agent-design}

We implement a simple web agent with chain-of-thought prompting~\citep{wei22cot}, and evaluate its performance across two axes: (1) the underlying LLM, and (2) the use of BrowserGym features. To study the effect of different design choices in our agents, we use flags to activate or deactivate certain features, such as \texttt{use\_thinking} for chain-of-thought.

\paragraphtight{Observation space:} Our observation space is composed of the goal, the current page's HTML and/or AXTree,\footnote{On WebArena and WorkArena we only use AXTrees because HTML is prohibitively large. On MiniWoB we use both AXTree and HTML as it consistently gave the best performance.} the currently focused element, and the error from the previous action if any. We activate / deactivate these features with flags \texttt{use\_last\_error} and \texttt{use\_focused\_element}, and we decide whether or not to augment each element with extra information using flags \texttt{coords=center} or \texttt{coords=box} for screen coordinates, and \texttt{extract\_visible\_tag} and \texttt{extract\_clickable\_tag} for whether elements are visible and/or clickable. Last, the flag \texttt{filter\_visible\_only} allow us to only include elements that are visible on the screen.

\paragraphtight{Action space:} We leverage the flexibility of BrowserGym with a \texttt{multi\_action} flag, which either allows agents to execute multiple actions per step (e.g., \texttt{click(12) click(52)}) or not. The \texttt{action\_set=bid} flag only permits primitives that use element identifiers, while \texttt{action\_set=bid+coord} also includes primitives operating with 2D coordinates \texttt{(x,y)}. This is useful in MiniWoB, where certain tasks require clicking at specific positions in SVG images. Finally, we also experiment with the textual description of the action space which can either include a short or long description for each primitive (\texttt{long\_description}), as well as examples of valid function calls or not (\texttt{individual\_examples}).

\paragraphtight{History:} To extend the horizon window of our agents, we experiment with \texttt{use\_action\_history} which includes in the prompt the history of actions since the beginning of the episode, \texttt{use\_error\_history} which includes all previous error messages, and \texttt{use\_think\_history} which re-injects the previous chain-of-though outputs into the prompt, thus creating an artificial memory that gives agents a chance to recall their previous thoughs.

\paragraphtight{Zero-shot examples:} In the prompt, we provide a single generic example of how the chain of thought and action outputs should be formatted. This contrasts with other methods~\citep{kim2023rciagent} where task-specific few-shot examples are provided, yet aligns with our objective of developing zero-shot agents able to solve a large range of new tasks.

\paragraphtight{Parse and retry:} Once the LLM provides an answer, we have a parsing loop that can re-prompt the agent up to 4 times to make it aware of a parsing mistake. This can save the agent from making basic mistakes and is mainly useful for less capable LLMs such as GPT-3.5. Once parsed, the action is executed via BrowserGym, which moves to the next step.

\paragraphtight{Language models:} Our study distinguishes between closed- and open-source LLMs. For the closed-source segment, we evaluate \textbf{GPT-3.5} (\texttt{gpt-3.5-turbo-1106}, 16K context) and \textbf{GPT-4o}~\citep{OpenAI2023GPT4TR} (\texttt{gpt-4o-2024-05-13}, 128K context), through OpenAI's API. In the realm of open-source LLMs, we sought a model that 1) understands code and HTML, 2) can manage a substantial context size, and 3) is instruction-finetuned. Our choice fell on \textbf{Llama3-70b}~\citep{meta2024llama3} (\texttt{meta-llama-3-70B-instruct}, 8K context), a recently released model that exhibits performances close to GPT-4. This model was deployed using Hugging Face's Text Generation Inference (TGI) library on 4 A100 GPUs. We also explore the effect of providing the screenshot of the page using GPT-4o vision and Set-of-Mark~\citep{yang2023set} as proposed in WebVoyager~\citep{he2024webvoyager}.

\paragraphtight{Prompt truncation:} We use a maximum prompt length of 40K tokens for GPT-4o, 15K for GPT-3.5 and 8K for Llama3. When the prompt is too large (\cref{fig:token_stats}), we progressively truncate the HTML and AXTree from the end until it fits the maximum allowed number of tokens.

\newcommand{\gpm}[1]{\textcolor{gray}{\tiny$\pm$#1}}

\newcolumntype{L}{>{\raggedright\arraybackslash}X}
\newcolumntype{R}{>{\raggedleft\arraybackslash}X}

\begin{table}[t] 
\caption{Success rate\gpm{Standard error} (SR \gpm{SE}) of all agents on MiniWoB, WorkArena, and WebArena. Bolded numbers represent the average success rate over the entire corresponding benchmark.}
\noindent\resizebox{\columnwidth}{!}{ 
\begin{tabular}{l r@{\hspace{2pt}}l r@{\hspace{2pt}}l r@{\hspace{2pt}}l r@{\hspace{2pt}}l r@{\hspace{2pt}}l}
\toprule
\textbf{Task Category} & \multicolumn{2}{c}{\textbf{GPT-4o}} & \multicolumn{2}{c}{\textbf{GPT-4o-V}} & \multicolumn{2}{c}{\textbf{GPT-3.5}} & \multicolumn{2}{c}{\textbf{Llama3}} \\
 & \textbf{SR \%} & \gpm{SE} & \textbf{SR \%} & \gpm{SE} & \textbf{SR \%} & \gpm{SE} & \textbf{SR \%} & \gpm{SE} \\
\midrule
\textbf{WorkArena} {\tiny(33 tasks)} & \textbf{42.7} & \gpm{1.5} & \textbf{41.8} & \gpm{1.7} & \textbf{6.1} & \gpm{1.3} & \textbf{17.9} & \gpm{1.5} \\
\quad Dashboard {\tiny(4)} & 62.5 & \gpm{6.8} & 72.5 & \gpm{6.0} & 20.0 & \gpm{4.8} & 37.5 & \gpm{6.0} \\
\quad Form {\tiny(5)} & 40.0 & \gpm{5.9} & 34.0 & \gpm{4.8} & 2.0 & \gpm{2.5} & 32.0 & \gpm{4.6} \\
\quad Knowledge {\tiny(1)} & 80.0 & \gpm{12.2} & 70.0 & \gpm{13.9} & 0.0 & \gpm{4.3} & 30.0 & \gpm{12.3} \\
\quad List-filter {\tiny(6)} & 0.0 & \gpm{1.6} & 0.0 & \gpm{1.7} & 0.0 & \gpm{1.6} & 0.0 & \gpm{1.8} \\
\quad List-sort {\tiny(6)} & 10.0 & \gpm{3.8} & 13.3 & \gpm{4.0} & 8.3 & \gpm{3.7} & 1.7 & \gpm{2.5} \\
\quad Menu {\tiny(2)} & 60.0 & \gpm{8.0} & 90.0 & \gpm{6.0} & 5.0 & \gpm{4.7} & 0.0 & \gpm{2.9} \\
\quad Service catalog {\tiny(9)} & 77.8 & \gpm{3.2} & 65.6 & \gpm{3.6} & 5.6 & \gpm{2.3} & 26.7 & \gpm{3.4} \\
\midrule
\textbf{MiniWoB} {\tiny(125 tasks)} & \textbf{66.1} & \gpm{1.0} & \textbf{67.7} & \gpm{1.0} & \textbf{38.9} & \gpm{1.1} & \textbf{62.6} & \gpm{0.6} \\
\quad WebGum Subset {\tiny(56)} & 82.9 & \gpm{1.5} & 83.2 & \gpm{1.5} & 53.6 & \gpm{1.4} & 80.5 & \gpm{1.0} \\
\midrule
\textbf{WebArena} {\tiny(812 tasks)} & \textbf{23.5} & \gpm{0.7} &\textbf{24.0} & \gpm{0.6} &  \textbf{6.7} & \gpm{0.6} & \textbf{11.0} & \gpm{0.6} \\
\quad Content-and-config {\tiny(411)} & 25.8 & \gpm{1.0} & 26.8 & \gpm{0.9} & 8.8 & \gpm{0.8} & 12.7 & \gpm{0.9} \\
\quad Information-seeking {\tiny(325)} & 22.5 & \gpm{1.0} & 22.5 & \gpm{0.9} & 4.3 & \gpm{0.9} & 9.8 & \gpm{1.1} \\
\quad Navigation {\tiny(76)} & 15.8 & \gpm{2.2} & 15.8 & \gpm{1.8} & 5.3 & \gpm{1.9} & 6.6 & \gpm{1.9} \\
\bottomrule
\end{tabular}
}
\label{tab:acc-summary}
\end{table}

\subsection{Experimental Protocol} \label{sec:protocol}

\paragraphtight{Standard Error:} To be able to run a range of experiments under a fixed budget, we limit the number of seeds to 10 per task for MiniWoB and WorkArena, and 1 per task for WebArena. After averaging results over each benchmark (or subset of it), we usually observe a sufficiently low standard error to draw the needed conclusions. We use stratified bootstrap
to obtain 1,000 samples of the mean and report the average and standard deviation of these means as success rate and standard error.

\paragraphtight{Max step:} In all tasks we give our agents a maximum of 15 steps per episode, as in \citet{he2024webvoyager}. This ensures a low-performing agent will not wander for too long if they are incapable of solving the task. Note that 15 steps is considered sufficient for MiniWoB, but on WorkArena some tasks might require more than 15 steps to complete unless the agent runs in multi-action mode (see \cref{app:wa-tasks} \cref{tab:workarena_all_tasks} for an upper bound).

\paragraphtight{Model selection:} We evaluate all LLMs using a common agent code-base that we tune by activating / deactivating flags (see \cref{sec:agent-design}). For each LLM we find the best agent configuration via random search on MiniWoB and WorkArena. We then fix the final configuration (one per LLM), and re-run a full evaluation on all benchmarks with a different seed.\footnote{WebArena being deterministic, we don't use it for tuning.} Final configurations are reported in \cref{tab:best-agent-configs}.

\subsection{Results}

We report the final performance of our agents on MiniWoB~\citep{LiuL18MiniWoB}, WebArena~\citep{zhou2023webarena} and WorkArena in Table~\ref{tab:acc-summary}. We emphasize our key findings below.

\paragraphtight{GPT-4o's Superiority:} The data unequivocally demonstrate GPT-4o's dominance over GPT-3.5 and Llama3, on all benchmarks.  The performance gap is particularly striking on WorkArena, where GPT-4o achieves 43\% success, in stark contrast to GPT-3.5's 6.1\% and Llama3's 17.9\%. The performance disparity between LLMs is significantly more pronounced in WorkArena and WebArena than in MiniWoB, aligning with findings on the emergent properties of AI systems~\citep{wei2022emergent}. As the complexity of task increases, the necessity for a more advanced LLM to achieve any score becomes apparent, with noticeable improvements in performance correlating with the enhancement of capabilities.

\paragraphtight{WorkArena poses a great challenge:} Consistent with our expectations, our newly proposed benchmark, WorkArena, proves to be a significant challenge for current LLMs. While the tasks remain simple at a high-level (navigating menus, filling forms) and mostly require general knowledge, the difficulty arises primarily from the use of complex user interfaces from real-world software environments, resulting in long contexts and non-trivial interactions. Consequently, all agents exhibit low performance levels, with no agent achieving 100\% success in one specific task. Notably, the list-based tasks which require interacting with a non-standard HTML widget (\cref{app:wa-task-ui} \cref{fig:wa-filterlist-ui}) prove to be the most challenging ones with a clear 0\% success rate for all LLMs, while being fairly simple for a human to complete.

\paragraphtight{State-of-the-art on MiniWoB and WebArena:} Our GPT-4o agent demonstrates notably high performance on MiniWoB, achieving significantly greater success compared to other agents. The outcomes on the full benchmark and the WebGum subset (82.9\%) surpass those of prior studies on zero-shot web agents \citep{assouel2023the,zeng2023agenttuning}, underscoring the effectiveness of both our agent design and the features offered by BrowserGym. On WebArena, while the score obtained by our GPT-4o agent is only 23.5\%, it constitutes the best zero-shot performance reported so far, way above the 14.4\% succes rate obtained  using GPT-4 in the original paper\citep{zhou2023webarena}.

\paragraphtight{Open-source LLMs:} While our Llama3 agent still performs poorer than a closed-source GPT-4o, its performance is remarkably high compared to a GPT-3.5 agent despite its shorter context window (8K vs 16K). Most notably, in preliminary experiments with a Llama2 model we could not obtain any success (flat 0\%) on both WebArena and WorkArena, which makes us hopeful for the future of open-source LLMs. These results also indicate that web agent benchmarks, and WorkArena in particular, offer a great tool for evaluating emerging capabilities in LLMs, with larger performance gaps and more room for improvement compared to regular benchmarks \citep{llama2}.

\begin{table}[ht]
\caption{Ablation study for \textbf{GPT-4o} on MiniWoB and WorkArena. Success rate\gpm{Standard error} (SR \gpm{SE}) of all configurations. Each row modifies the initial configuration.}
\noindent\resizebox{\columnwidth}{!}{
\begin{tabular}{l r@{\hspace{2pt}}l r@{\hspace{2pt}}l}
\toprule
\multicolumn{1}{l}{\textbf{Configuration}} & \multicolumn{2}{c}{\textbf{MiniWoB}} & \multicolumn{2}{c}{\textbf{WorkArena}} \\
 & \textbf{SR \%} & \gpm{SE} & \textbf{SR \%} & \gpm{SE} \\
\midrule
Initial configuration & 68.2 & \gpm{1.0} & 45.5 & \gpm{2.2} \\
+multi\_actions & 68.5 & \gpm{1.0} & 40.6 & \gpm{2.0} \\
+coords=box,action\_set=bid+coord & 72.6 & \gpm{1.0} & 41.2 & \gpm{1.8} \\
+use\_think\_history & 66.7 & \gpm{0.9} & 42.4 & \gpm{2.3} \\
+use\_error\_history & 67.2 & \gpm{0.9} & 43.6 & \gpm{2.1} \\
-extract\_visible\_tag & 68.8 & \gpm{1.0} & 43.0 & \gpm{2.2} \\
\bottomrule
\end{tabular}
}
\label{tab:ablation-gpt-4o}
\end{table}

\begin{table}[ht]
\caption{Ablation study for \textbf{GPT-3.5} on MiniWoB and WorkArena. Success rate\gpm{Standard error} (SR \gpm{SE}) of all configurations. Each row modifies the initial configuration.}
\noindent\resizebox{\columnwidth}{!}{
\begin{tabular}{l r@{\hspace{2pt}}l r@{\hspace{2pt}}l}
\toprule
\multicolumn{1}{l}{\textbf{Configuration}} & \multicolumn{2}{c}{\textbf{MiniWoB}} & \multicolumn{2}{c}{\textbf{WorkArena}} \\
 & \textbf{SR \%} & \gpm{SE} & \textbf{SR \%} & \gpm{SE} \\
\midrule
Initial configuration & 41.3 & \gpm{1.1} & 8.5 & \gpm{1.3} \\
-use\_thinking & 30.2 & \gpm{1.0} & 6.1 & \gpm{1.2} \\
-use\_action\_history & 34.1 & \gpm{1.1} & 5.2 & \gpm{1.3} \\
+use\_think\_history & 36.2 & \gpm{1.2} & 3.0 & \gpm{1.0} \\
+use\_error\_history & 37.9 & \gpm{1.0} & 7.0 & \gpm{1.4} \\
+multi\_actions & 41.4 & \gpm{1.1} & 9.4 & \gpm{1.5} \\
+only\_visible\_elements & 38.1 & \gpm{1.0} & 5.8 & \gpm{1.3} \\
+long\_description & 39.5 & \gpm{1.0} & 6.4 & \gpm{1.2} \\
-individual\_examples & 34.4 & \gpm{1.0} & 8.2 & \gpm{1.3} \\
+coords=center, action\_set=bid+coord & 40.8 & \gpm{1.1} & 4.5 & \gpm{1.1} \\
+coords=box, action\_set=bid+coord & 40.5 & \gpm{1.2} & 6.1 & \gpm{1.4} \\
+use\_focused\_element & 40.3 & \gpm{1.1} & 9.4 & \gpm{1.5} \\
-extract\_visible\_tag & 37.8 & \gpm{1.0} & 6.7 & \gpm{1.3} \\
+extract\_clickable\_tag & 37.9 & \gpm{1.1} & 7.6 & \gpm{1.2} \\
\bottomrule
\end{tabular}
}
\label{tab:ablation-gpt-3.5}
\end{table}

\begin{table}[ht]
\caption{Ablation study for \textbf{Llama3} on MiniWoB and WorkArena. Success rate\gpm{Standard error} (SR \gpm{SE}) of all configurations. Each row modifies the initial configuration.}
\begin{tabular}{l r@{\hspace{2pt}}l r@{\hspace{2pt}}l}
\toprule
\multicolumn{1}{l}{\textbf{Configuration}} & \multicolumn{2}{c}{\textbf{MiniWoB}} & \multicolumn{2}{c}{\textbf{WorkArena}} \\
 & \textbf{SR \%} & \gpm{SE} & \textbf{SR \%} & \gpm{SE} \\
\midrule
Initial configuration & 59.8 & \gpm{1.0} & 20.0 & \gpm{2.3} \\
-use\_thinking & 48.6 & \gpm{1.1} & 8.5 & \gpm{1.7} \\
-use\_action\_history & 49.9 & \gpm{1.1} & 8.5 & \gpm{1.7} \\
-use\_think\_history & 52.2 & \gpm{1.1} & 18.8 & \gpm{2.1} \\
+multi\_actions & 63.0 & \gpm{1.0} & 17.6 & \gpm{2.1} \\
+long\_description & 55.5 & \gpm{1.0} & 17.6 & \gpm{2.0} \\
+use\_last\_error & 60.2 & \gpm{1.1} & 19.4 & \gpm{1.8} \\
-extract\_visible\_tag & 61.3 & \gpm{0.9} & 15.8 & \gpm{2.0} \\
\bottomrule
\end{tabular}
\label{tab:ablation-llama3}
\end{table}

\subsection{Ablation Study}

We report in \cref{tab:ablation-gpt-4o,tab:ablation-gpt-3.5,tab:ablation-llama3} an ablation study of each LLM on MiniWoB and WorkArena, for the most important agent flags.\footnote{The ablation study was done with a different seed, hence numbers differ slightly from those in \cref{tab:acc-summary}.} We present our findings below.

\paragraphtight{Chain-of-though is crucial:} In all of our 3 agents, asking the LLM to directly produce the next best action without chain-of-though is very detrimental to performance. This is clearly seen in \cref{tab:ablation-llama3}, with a drop of 10 points for Llama3 on both MiniWoB and WorkArena. This conclusion holds for all three LLMs very clearly.

\begin{figure}
    \centering
    \includegraphics[width=\linewidth]{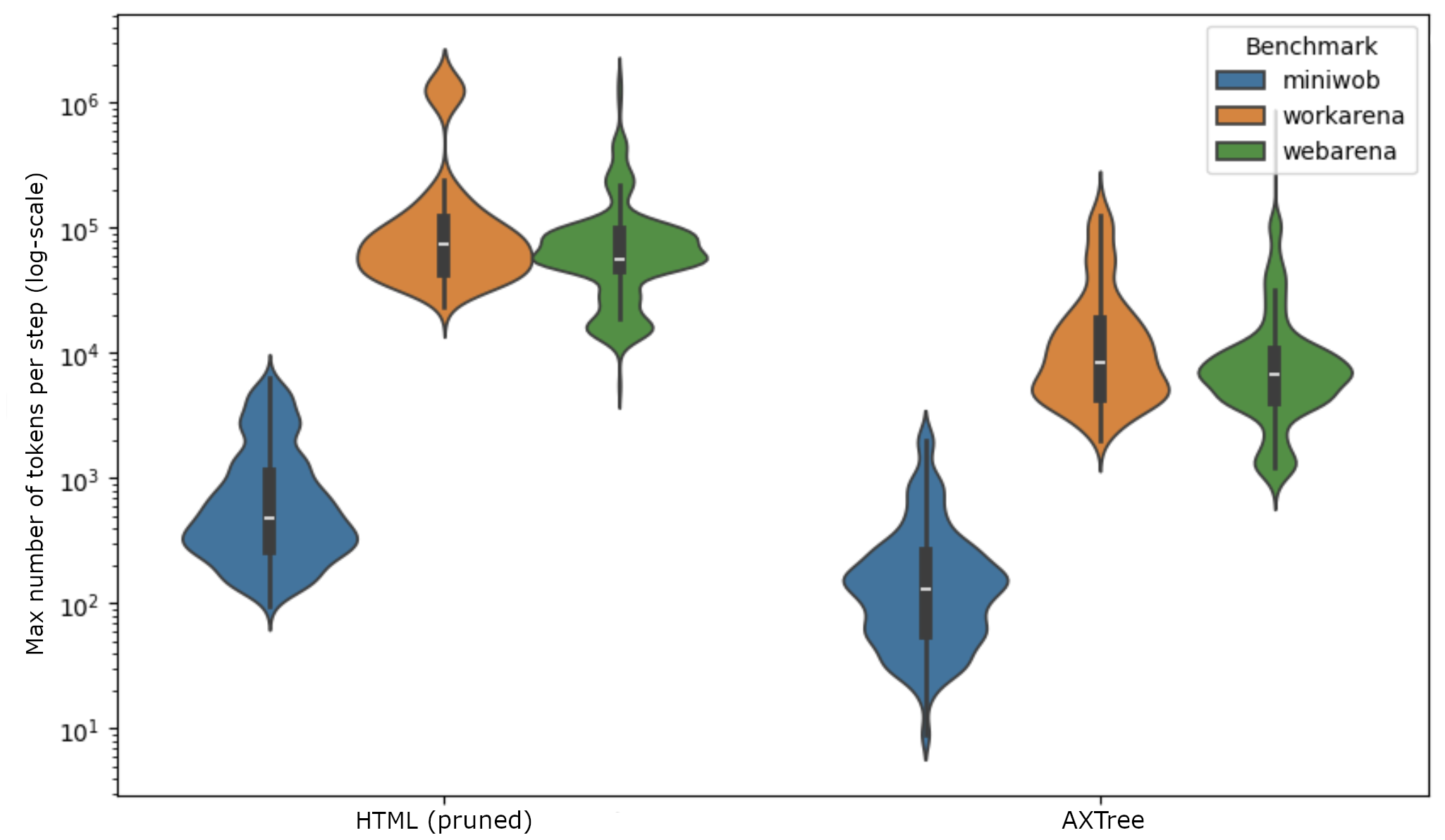}
    \caption{Comparative size analysis of different page observation modalities: HTML (left) and accessibility tree (right) across MiniWoB, WorkArena and WebArena.}
    \label{fig:token_stats}
\end{figure}

\paragraphtight{More is not always better:} As we gradually add more features, the prompt becomes longer and seems to overwhelm the LLM. This is mostly observed with GPT-3.5 and Llama3, the weaker LLMs, where seemingly harmless features such as a \texttt{long\_description} or \texttt{individual\_examples} in the action space description actually degrade performance. We hypothesise that since the AXTree is already large, adding too much non-crucial features is likely to distract the agent instead of helping it. We also note that in WorkArena, more features imply more prompt truncation for GPT-3.5 and Llama3 due to their limited context length (\cref{fig:token_stats}).

\paragraphtight{Memory can hurt:} The \texttt{use\_think\_history} flag provides an interesting capacity to our agents, that of remembering their thoughts from previous time steps. We observed in practice that this features works, in the sense that the agents do re-use information about previous pages. However, such a feature is mostly useful when the task requires some form of memory, such as navigating a first page to find some information (e.g., an address), and then going to another page (e.g., Google maps) to type this information. On both WorkArena and MiniWoB the vast majority of tasks does not require navigation, and all the necessary information is always present on the current page. What we observe instead, is that agents with this feature activated tend to stick to decisions decided in early time steps, even erroneous ones, and are somehow less prone to self-correction. This is mostly observed with the GPT agents, \cref{tab:ablation-gpt-4o,tab:ablation-gpt-3.5}.

\paragraphtight{Deceiving multimodal performance:} We report in \cref{tab:acc-summary} a complimentary experiment with GPT-4o-V, a vision-augmented GPT-4o model, to assess the capacity of multi-modal models at utilizing the screenshot of the current web page for solving web tasks. We augment the screenshot with Set-of-Mark~\citep{yang2023set} as proposed in WebVoyager~\citep{he2024webvoyager}, and we keep the rest of the flags identical to the GPT-4o best performing agent. The results are somewhat deceiving, with very minor performance improvements on MiniWoB and WebArena. Similarly poor multimodal performances are reported by \citet{xie2024osworld}, which could be explained by the lack of screen-related training data in current vision-language models. This leaves room for future development in this area.

\paragraphtight{2D features do not always help:} Several MiniWoB tasks require precise 2D understanding of the UI and interaction with $x,y$ coordinates. While we observe a notable performance gain on this benchmark when giving GPT-4o additional coordinate features and actions (\cref{tab:ablation-gpt-4o}), this does not translate to WorkArena where the tasks seem best solved using \texttt{bid} only. This seems to indicate that, while 2D actions are clearly required in specific cases (i.e., drawing on a web whiteboard), most web interactions actually do not require it, which can greatly simplify the agent design.

\section{Conclusion}

In this work, we introduced WorkArena, a new benchmark for the evaluation of web agents on tasks inspired by the day-to-day workflow of knowledge workers in the ServiceNow platform. WorkArena underscores the challenge of navigating real-world enterprise websites, notably dealing with large observations and complex, non-standard interfaces.

We also introduced BrowserGym, a robust, general-purpose environment for automated web agents, which encompasses an extensive list of features previously proposed in the literature (HTML, AXTree, screenshot, set-of-mark, code and high-level action space), as well as novel capabilities such as an interactive chat. BrowserGym allows for the flawless evaluation of web agents on multiple benchmarks within the same unified framework, and currently supports MiniWoB, WebArena and WorkArena.
Both contributions are open source and meant to serve as catalysts to accelerate the development of new web agents and their evaluation in terms of capability and potential impact on the real world.

We presented an empirical evaluation of GPT-3.5, GPT-4o and Llama3 -- among the most advanced general-purpose Large Language Models (LLMs) currently available for intruction-following and coding.
Specifically, we investigated their generalization performance as web agents in MiniWoB, WebArena and WorkArena.
Our results validate WorkArena as an unsolved, challenging benchmark that requires advanced reasoning capabilities over long contexts (HTML or AXTree), which seem to emerge only in very large models.

In future work we plan to integrate additional standard benchmarks into BrowserGym, such as WebShop~\citep{yao2022webshop} and WebVoyager~\citep{he2024webvoyager}. 
We also plan to expand WorkArena significantly by creating compositional tasks built upon the present benchmark, to obtain realistic complex workflows that require additional skills like retrieval, memorization, visual perception, and advanced reasoning.

Due to the multi-modal nature of web observations (textual and visual), and the potentially unlimited complexity of web tasks that can be designed (from toy setups like MiniWoB to harder benchmarks like WebArena and WorkArena), we believe that browser-based task automation provides the perfect testbed to evaluate the emergent capabilities of multimodal large language models. 
We hope that the present work will stimulate further progress in the community.

\section*{Impact Statement}

This research presents contributions aimed at facilitating the development of UI assistants, with a particular focus on browser-based web agents and their application in the workspace.
The emergence of such agents is bound to disrupt the way humans interact with digital software, and the way workers perform digital tasks.
Below, we reflect on the potential positive and negative societal impacts of this work.

\paragraphtight{Positive impacts}
\begin{itemize}
    \item \textbf{Productivity.} A prominent benefit of UI assistants / web agents is the significant boost in worker productivity that would be achieved by automating repetitive and monotonous tasks. This automation would not only streamline digital workflows but also free up valuable time for employees to engage in complex problem-solving and creative tasks, thereby enhancing overall work quality and innovation.
    \item \textbf{Accessibility} Further, UI assistants / web agents could lead to a substantial leap in digital accessibility, opening new employment opportunities for individuals who may have previously been excluded from certain roles due to disabilities, such as visual impairments. This advancement could also mitigate labor shortages by expanding the talent pool, benefiting both employers and workers in the job market.
\end{itemize}

\paragraphtight{Negative impacts} 
\begin{itemize}
    \item \textbf{Labor displacement.} One concern is job market disruption, including labor displacement. However, we believe this disruption is more likely to lead to the evolution of job roles rather than their outright replacement. A prime example can be observed in the translation industry, which has been significantly disrupted by automatic translation~\citet{VanDerMeer2021}, but continues to thrive nonetheless. Interestingly, benchmarks such as WorkArena may help in forecasting which job roles are more likely to be affected by automation and help take preventive measures to minimize downsides (e.g., preemptive human reskilling).
    \item \textbf{Cybersecurity.} On another note, human-like web agents the potential for increased and elaborate cyberattacks via agents mimicking human interactions. This necessitates the implementation of security measures, such as the use of constrained language models (e.g., GPT-4~\citep{OpenAI2023GPT4TR}).
    \item \textbf{Privacy.} The deployment of LLM-based web agents in the workplace also raises privacy concerns due to the necessity of transmitting sensitive information, requiring further security research.
    \item \textbf{Environment.} Additionally, the significant energy consumption associated with the extensive use of LLMs for inference presents environmental challenges that must be addressed before their widespread implementation.
\end{itemize}

Web agents, as a disruptive technology, introduce several potential negative impacts. We believe that all of the above points warrant careful consideration by the research community.

\bibliography{biblio}

\begin{thebibliography}{37}
\providecommand{\natexlab}[1]{#1}
\providecommand{\url}[1]{\texttt{#1}}
\expandafter\ifx\csname urlstyle\endcsname\relax
  \providecommand{\doi}[1]{doi: #1}\else
  \providecommand{\doi}{doi: \begingroup \urlstyle{rm}\Url}\fi

\bibitem[Assouel et~al.(2023)Assouel, Marty, Caccia, Laradji, Drouin, Rajeswar, Palacios, Cappart, Vazquez, Chapados, Gasse, and Lacoste]{assouel2023the}
Assouel, R., Marty, T., Caccia, M., Laradji, I., Drouin, A., Rajeswar, S., Palacios, H., Cappart, Q., Vazquez, D., Chapados, N., Gasse, M., and Lacoste, A.
\newblock The unsolved challenges of {LLM}s in open-ended web tasks: A case study.
\newblock In \emph{NeurIPS 2023 Foundation Models for Decision Making Workshop}, 2023.
\newblock URL \url{https://openreview.net/forum?id=jt3il4fC5B}.

\bibitem[Brockman et~al.(2016)Brockman, Cheung, Pettersson, Schneider, Schulman, Tang, and Zaremba]{software/gym}
Brockman, G., Cheung, V., Pettersson, L., Schneider, J., Schulman, J., Tang, J., and Zaremba, W.
\newblock {OpenAI} gym, 2016.

\bibitem[Deng et~al.(2023)Deng, Gu, Zheng, Chen, Stevens, Wang, Sun, and Su]{deng2023mind2web}
Deng, X., Gu, Y., Zheng, B., Chen, S., Stevens, S., Wang, B., Sun, H., and Su, Y.
\newblock {Mind2Web}: Towards a generalist agent for the web.
\newblock \emph{arXiv}, abs/2306.06070, 2023.

\bibitem[Furuta et~al.(2023)Furuta, Nachum, Lee, Matsuo, Gu, and Gur]{furuta2023multimodal}
Furuta, H., Nachum, O., Lee, K.-H., Matsuo, Y., Gu, S.~S., and Gur, I.
\newblock Multimodal web navigation with instruction-finetuned foundation models.
\newblock \emph{arXiv}, abs/2305.11854, 2023.
\newblock URL \url{https://arxiv.org/abs/2305.11854}.

\bibitem[{Google}(2023)]{software/chromedevtools}
{Google}.
\newblock Chrome devtools protocol, 2023.
\newblock URL \url{https://chromedevtools.github.io/devtools-protocol/}.

\bibitem[Gur et~al.(2023{\natexlab{a}})Gur, Furuta, Huang, Safdari, Matsuo, Eck, and Faust]{gur2023real}
Gur, I., Furuta, H., Huang, A., Safdari, M., Matsuo, Y., Eck, D., and Faust, A.
\newblock A real-world webagent with planning, long context understanding, and program synthesis.
\newblock \emph{arXiv preprint arXiv:2307.12856}, 2023{\natexlab{a}}.

\bibitem[Gur et~al.(2023{\natexlab{b}})Gur, Furuta, Huang, Safdari, Matsuo, Eck, and Faust]{gur2023webagent}
Gur, I., Furuta, H., Huang, A., Safdari, M., Matsuo, Y., Eck, D., and Faust, A.
\newblock A real-world {WebAgent} with planning, long context understanding, and program synthesis.
\newblock \emph{arXiv}, abs/2307.12856, 2023{\natexlab{b}}.
\newblock URL \url{https://arxiv.org/abs/2307.12856}.

\bibitem[He et~al.(2024)He, Yao, Ma, Yu, Dai, Zhang, Lan, and Yu]{he2024webvoyager}
He, H., Yao, W., Ma, K., Yu, W., Dai, Y., Zhang, H., Lan, Z., and Yu, D.
\newblock {WebVoyager}: Building an end-to-end web agent with large multimodal models.
\newblock \emph{arXiv}, abs/2401.13919, 2024.
\newblock URL \url{https://arxiv.org/abs/2401.13919}.

\bibitem[Huang et~al.(2024)Huang, Li, Li, and Li]{macromining}
Huang, F., Li, G., Li, T., and Li, Y.
\newblock Automatic macro mining from interaction traces at scale.
\newblock In \emph{Proceedings of the CHI Conference on Human Factors in Computing Systems}, CHI '24, New York, NY, USA, 2024. Association for Computing Machinery.
\newblock ISBN 9798400703300.
\newblock \doi{10.1145/3613904.3642074}.
\newblock URL \url{https://doi.org/10.1145/3613904.3642074}.

\bibitem[Humphreys et~al.(2022)Humphreys, Raposo, Pohlen, Thornton, Chhaparia, Muldal, Abramson, Georgiev, Santoro, and Lillicrap]{Humphreys22rl4MiniWoB}
Humphreys, P.~C., Raposo, D., Pohlen, T., Thornton, G., Chhaparia, R., Muldal, A., Abramson, J., Georgiev, P., Santoro, A., and Lillicrap, T.
\newblock A data-driven approach for learning to control computers.
\newblock In \emph{International Conference on Machine Learning (ICML)}, 2022.

\bibitem[Kim et~al.(2023)Kim, Baldi, and McAleer]{kim2023rciagent}
Kim, G., Baldi, P., and McAleer, S.
\newblock Language models can solve computer tasks.
\newblock \emph{arXiv}, abs/2303.17491, 2023.
\newblock URL \url{https://arxiv.org/abs/2303.17491}.

\bibitem[Li et~al.(2020)Li, He, Zhou, Zhang, and Baldridge]{seq2act}
Li, Y., He, J., Zhou, X., Zhang, Y., and Baldridge, J.
\newblock Mapping natural language instructions to mobile ui action sequences.
\newblock In \emph{Annual Conference of the Association for Computational Linguistics (ACL 2020)}, 2020.
\newblock URL \url{https://www.aclweb.org/anthology/2020.acl-main.729.pdf}.

\bibitem[Liu et~al.(2018)Liu, Guu, Pasupat, Shi, and Liang]{LiuL18MiniWoB}
Liu, E.~Z., Guu, K., Pasupat, P., Shi, T., and Liang, P.
\newblock Reinforcement learning on web interfaces using workflow-guided exploration.
\newblock In \emph{International Conference on Learning Representations (ICLR)}, 2018.

\bibitem[Liu et~al.(2023{\natexlab{a}})Liu, Yu, Zhang, Xu, Lei, Lai, Gu, Ding, Men, Yang, Zhang, Deng, Zeng, Du, Zhang, Shen, Zhang, Su, Sun, Huang, Dong, and Tang]{liu2023agentbench}
Liu, X., Yu, H., Zhang, H., Xu, Y., Lei, X., Lai, H., Gu, Y., Ding, H., Men, K., Yang, K., Zhang, S., Deng, X., Zeng, A., Du, Z., Zhang, C., Shen, S., Zhang, T., Su, Y., Sun, H., Huang, M., Dong, Y., and Tang, J.
\newblock {AgentBench}: Evaluating {LLMs} as agents.
\newblock \emph{arXiv}, abs/2308.03688, 2023{\natexlab{a}}.
\newblock URL \url{https://arxiv.org/abs/2308.03688}.

\bibitem[Liu et~al.(2023{\natexlab{b}})Liu, Yao, Zhang, Xue, Heinecke, Murthy, Feng, Chen, Niebles, Arpit, Xu, Mui, Wang, Xiong, and Savarese]{liu2023bolaa}
Liu, Z., Yao, W., Zhang, J., Xue, L., Heinecke, S., Murthy, R., Feng, Y., Chen, Z., Niebles, J.~C., Arpit, D., Xu, R., Mui, P., Wang, H., Xiong, C., and Savarese, S.
\newblock {BOLAA}: Benchmarking and orchestrating {LLM}-augmented autonomous agents.
\newblock \emph{arXiv}, abs/2308.05960, 2023{\natexlab{b}}.

\bibitem[L{\`u} et~al.(2024)L{\`u}, Kasner, and Reddy]{lu2024weblinx}
L{\`u}, X.~H., Kasner, Z., and Reddy, S.
\newblock Weblinx: Real-world website navigation with multi-turn dialogue.
\newblock \emph{arXiv preprint arXiv:2402.05930}, 2024.

\bibitem[Maas(2020)]{maas2020knowledge}
Maas, M.
\newblock {Knowledge 2020: ``The digital workflow revolution has just begun''}.
\newblock Technical report, Sprinklr, 2020.
\newblock URL \url{https://www.linkedin.com/pulse/knowledge-2020-digital-workflow-revolution-has-just-begun-maas/}.

\bibitem[Mastantuono(2023)]{fortune500_2023}
Mastantuono, G.
\newblock {ServiceNow joins the prestigious Fortune 500 list}.
\newblock \url{https://www.servicenow.com/blogs/2023/servicenow-joins-fortune-500-list.html}, 2023.
\newblock Accessed: 2024-01-29.

\bibitem[Meta(2024)]{meta2024llama3}
Meta.
\newblock Llama 3: Meta's latest large language model.
\newblock \url{https://github.com/meta-llama/llama3}, 2024.
\newblock Accessed: 2024-06-03.

\bibitem[{Microsoft}(2023)]{Playwright}
{Microsoft}.
\newblock Playwright for {P}ython documentation, 2023.
\newblock URL \url{https://playwright.dev/python/}.

\bibitem[Nakano et~al.(2021)Nakano, Hilton, Balaji, Wu, Ouyang, Kim, Hesse, Jain, Kosaraju, Saunders, Jiang, Cobbe, Eloundou, Krueger, Button, Knight, Chess, and Schulman]{nakano2021webgpt}
Nakano, R., Hilton, J., Balaji, S., Wu, J., Ouyang, L., Kim, C., Hesse, C., Jain, S., Kosaraju, V., Saunders, W., Jiang, X., Cobbe, K., Eloundou, T., Krueger, G., Button, K., Knight, M., Chess, B., and Schulman, J.
\newblock {WebGPT}: Browser-assisted question-answering with human feedback.
\newblock \emph{arXiv}, abs/2112.09332, 2021.
\newblock URL \url{https://arxiv.org/abs/2112.09332}.

\bibitem[OpenAI(2023)]{OpenAI2023GPT4TR}
OpenAI.
\newblock {GPT-4} technical report.
\newblock \emph{ArXiv}, abs/2303.08774, 2023.
\newblock URL \url{https://arxiv.org/abs/2303.08774}.

\bibitem[Rawles et~al.(2023)Rawles, Li, Rodriguez, Riva, and Lillicrap]{aitw23}
Rawles, C., Li, A., Rodriguez, D., Riva, O., and Lillicrap, T.
\newblock Androidinthewild: A large-scale dataset for android device control.
\newblock In Oh, A., Naumann, T., Globerson, A., Saenko, K., Hardt, M., and Levine, S. (eds.), \emph{Advances in Neural Information Processing Systems}, volume~36, pp.\  59708--59728, 2023.

\bibitem[SAE(2021)]{sae/levels-automation}
SAE.
\newblock {Taxonomy and Definitions for Terms Related to Driving Automation Systems for On-Road Motor Vehicles}.
\newblock Technical report, Society of Automotive Engineers (SAE), 04 2021.
\newblock URL \url{https://doi.org/10.4271/J3016_202104}.

\bibitem[{ServiceNow}(2023)]{servicenow/vancouver}
{ServiceNow}.
\newblock Vancouver release notes.
\newblock Online, 2023.
\newblock Available at: \url{https://docs.servicenow.com/bundle/vancouver-release-notes/}.

\bibitem[Shi et~al.(2017{\natexlab{a}})Shi, Karpathy, Fan, Hernandez, and Liang]{Shi2017MiniWoB}
Shi, T., Karpathy, A., Fan, L., Hernandez, J., and Liang, P.
\newblock World of bits: An open-domain platform for web-based agents.
\newblock In \emph{International Conference on Machine Learning (ICML)}, 2017{\natexlab{a}}.

\bibitem[Shi et~al.(2017{\natexlab{b}})Shi, Karpathy, Fan, Hernandez, and Liang]{shi2017world}
Shi, T., Karpathy, A., Fan, L., Hernandez, J., and Liang, P.
\newblock World of bits: An open-domain platform for web-based agents.
\newblock \emph{ICML}, 2017{\natexlab{b}}.

\bibitem[Touvron et~al.(2023)Touvron, Martin, Stone, Albert, Almahairi, Babaei, Bashlykov, Batra, Bhargava, Bhosale, Bikel, Blecher, Canton-Ferrer, Chen, Cucurull, Esiobu, Fernandes, Fu, Fu, Fuller, Gao, Goswami, Goyal, Hartshorn, Hosseini, Hou, Inan, Kardas, Kerkez, Khabsa, Kloumann, Korenev, Koura, Lachaux, Lavril, Lee, Liskovich, Lu, Mao, Martinet, Mihaylov, Mishra, Molybog, Nie, Poulton, Reizenstein, Rungta, Saladi, Schelten, Silva, Smith, Subramanian, Tan, Tang, Taylor, Williams, Kuan, Xu, Yan, Zarov, Zhang, Fan, Kambadur, Narang, Rodriguez, Stojnic, Edunov, and Scialom]{llama2}
Touvron, H., Martin, L., Stone, K., Albert, P., Almahairi, A., Babaei, Y., Bashlykov, N., Batra, S., Bhargava, P., Bhosale, S., Bikel, D., Blecher, L., Canton-Ferrer, C., Chen, M., Cucurull, G., Esiobu, D., Fernandes, J., Fu, J., Fu, W., Fuller, B., Gao, C., Goswami, V., Goyal, N., Hartshorn, A., Hosseini, S., Hou, R., Inan, H., Kardas, M., Kerkez, V., Khabsa, M., Kloumann, I., Korenev, A., Koura, P.~S., Lachaux, M.-A., Lavril, T., Lee, J., Liskovich, D., Lu, Y., Mao, Y., Martinet, X., Mihaylov, T., Mishra, P., Molybog, I., Nie, Y., Poulton, A., Reizenstein, J., Rungta, R., Saladi, K., Schelten, A., Silva, R., Smith, E.~M., Subramanian, R., Tan, X.~E., Tang, B., Taylor, R., Williams, A., Kuan, J.~X., Xu, P., Yan, Z., Zarov, I., Zhang, Y., Fan, A., Kambadur, M., Narang, S., Rodriguez, A., Stojnic, R., Edunov, S., and Scialom, T.
\newblock Llama 2: Open foundation and fine-tuned chat models.
\newblock \emph{CoRR}, abs/2307.09288, 2023.

\bibitem[van~der Meer(2021)]{VanDerMeer2021}
van~der Meer, J.
\newblock A journey into the future of the translation industry, 2021.
\newblock URL \url{https://www.taus.net/resources/blog/a-journey-into-the-future-of-the-translation-industry}.
\newblock Accessed: 2024-02-01.

\bibitem[Wei et~al.(2022{\natexlab{a}})Wei, Tay, Bommasani, Raffel, Zoph, Borgeaud, Yogatama, Bosma, Zhou, Metzler, et~al.]{wei2022emergent}
Wei, J., Tay, Y., Bommasani, R., Raffel, C., Zoph, B., Borgeaud, S., Yogatama, D., Bosma, M., Zhou, D., Metzler, D., et~al.
\newblock Emergent abilities of large language models.
\newblock \emph{arXiv preprint arXiv:2206.07682}, 2022{\natexlab{a}}.

\bibitem[Wei et~al.(2022{\natexlab{b}})Wei, Wang, Schuurmans, Bosma, ichter, Xia, Chi, Le, and Zhou]{wei22cot}
Wei, J., Wang, X., Schuurmans, D., Bosma, M., ichter, b., Xia, F., Chi, E., Le, Q.~V., and Zhou, D.
\newblock Chain-of-thought prompting elicits reasoning in large language models.
\newblock In Koyejo, S., Mohamed, S., Agarwal, A., Belgrave, D., Cho, K., and Oh, A. (eds.), \emph{Advances in Neural Information Processing Systems}, volume~35, pp.\  24824--24837. Curran Associates, Inc., 2022{\natexlab{b}}.
\newblock URL \url{https://proceedings.neurips.cc/paper_files/paper/2022/file/9d5609613524ecf4f15af0f7b31abca4-Paper-Conference.pdf}.

\bibitem[Xie et~al.(2024)Xie, Zhang, Chen, Li, Zhao, Cao, Hua, Cheng, Shin, Lei, Liu, Xu, Zhou, Savarese, Xiong, Zhong, and Yu]{xie2024osworld}
Xie, T., Zhang, D., Chen, J., Li, X., Zhao, S., Cao, R., Hua, T.~J., Cheng, Z., Shin, D., Lei, F., Liu, Y., Xu, Y., Zhou, S., Savarese, S., Xiong, C., Zhong, V., and Yu, T.
\newblock Osworld: Benchmarking multimodal agents for open-ended tasks in real computer environments, 2024.

\bibitem[Yang et~al.(2023)Yang, Zhang, Li, Zou, Li, and Gao]{yang2023set}
Yang, J., Zhang, H., Li, F., Zou, X., Li, C., and Gao, J.
\newblock Set-of-mark prompting unleashes extraordinary visual grounding in gpt-4v.
\newblock \emph{arXiv preprint arXiv:2310.11441}, 2023.

\bibitem[Yao et~al.(2022)Yao, Chen, Yang, and Narasimhan]{yao2022webshop}
Yao, S., Chen, H., Yang, J., and Narasimhan, K.
\newblock {WebShop}: Towards scalable real-world web interaction with grounded language agents.
\newblock In \emph{Advances in Neural Information Processing Systems (NeurIPS)}, 2022.

\bibitem[Yao et~al.(2023)Yao, Zhao, Yu, Du, Shafran, Narasimhan, and Cao]{yao2023react}
Yao, S., Zhao, J., Yu, D., Du, N., Shafran, I., Narasimhan, K., and Cao, Y.
\newblock {ReAct}: Synergizing reasoning and acting in language models.
\newblock \emph{arXiv}, abs/2210.03629, 2023.
\newblock URL \url{https://arxiv.org/abs/2210.03629}.

\bibitem[Zeng et~al.(2023)Zeng, Liu, Lu, Wang, Liu, Dong, and Tang]{zeng2023agenttuning}
Zeng, A., Liu, M., Lu, R., Wang, B., Liu, X., Dong, Y., and Tang, J.
\newblock Agenttuning: Enabling generalized agent abilities for llms.
\newblock \emph{arXiv preprint arXiv:2310.12823}, 2023.

\bibitem[Zhou et~al.(2023)Zhou, Xu, Zhu, Zhou, Lo, Sridhar, Cheng, Bisk, Fried, Alon, and Neubig]{zhou2023webarena}
Zhou, S., Xu, F.~F., Zhu, H., Zhou, X., Lo, R., Sridhar, A., Cheng, X., Bisk, Y., Fried, D., Alon, U., and Neubig, G.
\newblock Webarena: A realistic web environment for building autonomous agents.
\newblock \emph{ArXiv}, abs/2307.13854, 2023.
\newblock URL \url{https://arxiv.org/abs/2307.13854}.

\end{thebibliography}
\bibliographystyle{icml2024}

\newpage
\appendix

\onecolumn

\section{WorkArena -- Additional Details}\label{app:workarena}

\subsection{Tasks} \label{app:wa-tasks}

This section provides additional details on each type of task included in the benchmark.

\begin{table}[htb]
    \centering
    \footnotesize
    \setlength{\tabcolsep}{12pt}
    \caption{List of all tasks available in WorkArena, grouped by category. \emph{Instances:} The number of instances corresponds to the number of instantiations of the parameters of the tasks (e.g., values to input into a specific field). Due to the combinatorial nature of list and form tasks, which resulted in an exceedingly large pool of potential instances, we chose to cap the number of instances at 1,000, selected randomly. \emph{Oracle Actions:} The number of Playwright actions required by the oracle functions (see \cref{sec:browsergym}) to solve the tasks. This is indicative of task complexity, but we emphasize that human-coded oracle functions, while correct, might not correspond to optimal solutions. Results are averaged over 10 randomly sampled instances and shown along with the standard deviation.}
    \label{tab:workarena_all_tasks}
    \begin{tabular}{@{}cMp{1in}c@{}}
        \toprule
        {\bf Category} & {\bf Task Name} & {\bf Instances} & {\bf Oracle Actions} \\
        \midrule
        \multicolumn{1}{c}{\multirow{12}{*}{\makecell{Lists\\(12 tasks)}}} & FilterAssetList & 1,000 & 17.3 $\pm$ 6.5\\
        & FilterChangeRequestList & 1,000 & 18.7 $\pm$ 4.7\\
        & FilterHardwareList & 1,000 & 18.4 $\pm$ 5.6\\
        & FilterIncidentList & 1,000 & 16.2 $\pm$ 4.2\\
        & FilterServiceCatalogItemList & 1,000 & 19.9 $\pm$ 5.9\\
        & FilterUserList & 1,000 & 12.7 $\pm$ 3.2\\
        & SortAssetList & 150 & 7.4 $\pm$ 2.3\\
        & SortChangeRequestList & 150 & 7.7 $\pm$ 1.6\\
        & SortHardwareList & 150 & 8 $\pm$ 2.3\\
        & SortIncidentList & 150 & 8 $\pm$ 2.7\\
        & SortServiceCatalogItemList & 150 & 8.3 $\pm$ 2.5\\
        & SortUserList & 150 & 7.7 $\pm$ 2.1\\
        \midrule
        \multicolumn{1}{c}{\multirow{5}{*}{\makecell{Forms\\(5 tasks)}}} & CreateChangeRequest & 1,000 & 21.5 $\pm$ 6.2\\
        & CreateIncident & 1,000 & 23 $\pm$ 7.9\\
        & CreateHardwareAsset & 1,000 & 47.1 $\pm$ 10.9\\
        & CreateProblem & 1,000 & 10 $\pm$ 3.4\\
        & CreateUser & 1,000 & 17.9 $\pm$ 5.2\\
        \midrule
        \multicolumn{1}{c}{\multirow{1}{*}{\makecell{Knowledge Bases (1 task)}}} & KnowledgeBaseSearch &  1,000 & 4.0 $\pm$ 0.0\\
        \midrule
        \multicolumn{1}{c}{\multirow{9}{*}{\makecell{Service Catalogs\\(9 tasks)}}} & OrderDeveloperLaptopMac & 1,000 & 8.7 $\pm$ 0.9\\
        & OrderIpadMini & 80 & 6.0 $\pm$ 0.0\\
        & OrderIpadPro & 60 & 6.0 $\pm$ 0.0\\
        & OrderSalesLaptop & 1,000 & 9.0 $\pm$ 0.8\\
        & OrderStandardLaptop & 1,000 & 8.0 $\pm$ 0.6\\
        & OrderAppleWatch & 10 & 4.0 $\pm$ 0.0\\
        & OrderAppleMacBookPro15 & 10 & 4.0 $\pm$ 0.0\\
        & OrderDevelopmentLaptopPC & 40 & 6.0 $\pm$ 0.0\\
        & OrderLoanerLaptop & 350 & 8.0 $\pm$ 0.0\\
        \midrule
        \multicolumn{1}{c}{\multirow{2}{*}{\makecell{Menus\\(2 tasks)}}} & AllMenu & 1,000 & 3.0 $\pm$ 0.0\\
        & Impersonation & 600 & 7.0 $\pm$ 0.0\\
        \midrule
        \multicolumn{1}{c}{\multirow{2}{*}{\makecell{Dashboards\\(4 tasks)}}} & SingleChartValueRetrieval & 1000 & 1.0 $\pm$ 0.0\\
        & SingleChartMinMaxRetrieval & 346 &  1.0 $\pm$ 0.0\\
        & MultiChartValueRetrieval & 444 &  2.0 $\pm$ 0.0\\
        & MultiChartMinMaxRetrieval & 72 &  2.0 $\pm$ 0.0\\
        \midrule
        \textbf{Total (33 tasks)} &   & \textbf{\wainstancecount}\\
        \bottomrule
    \end{tabular}
\end{table}

\subsection{Task User Interface Examples}\label{app:wa-task-ui}

\begin{figure*}
    \centering
    \includegraphics[width=0.95\linewidth]{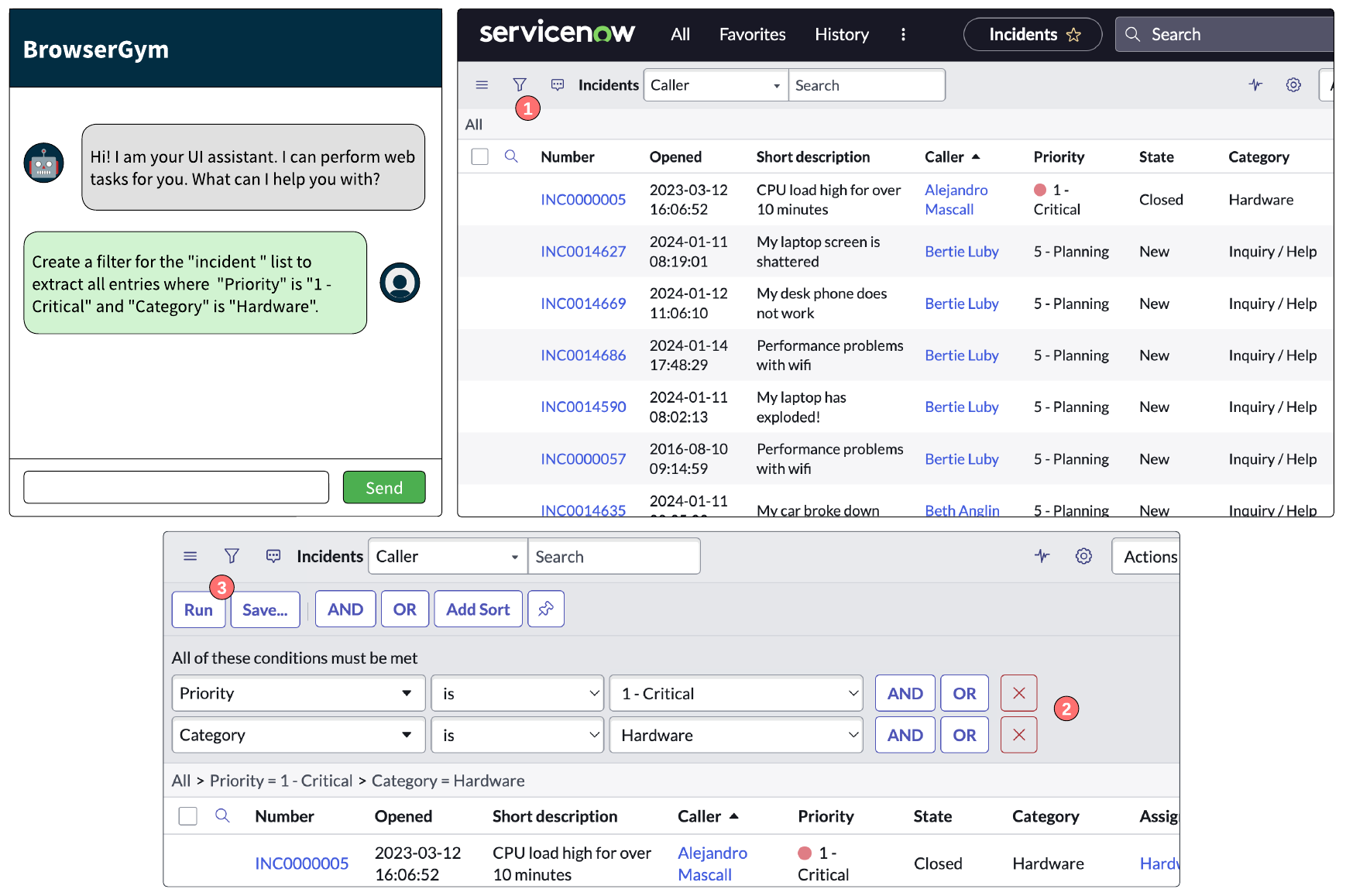}
    \caption{Example ``FilterIncidentList" Task -- The goal is given to the agent in natural language. As can be seen, the goal is designed to be very explicit, leaving no ambiguity on the task to perform. Here, the agent must expose the filter creation menu, by clicking on the appropriate icon~\circledigit{1}. Then, it must add conditions one by one and fill them out accordingly~\circledigit{2}. Finally, it must apply the filter using the ``Run'' button~\circledigit{3}.}
    \label{fig:wa-filterlist-ui}
\end{figure*}

\begin{figure*}
    \centering
    \includegraphics[width=0.95\linewidth]{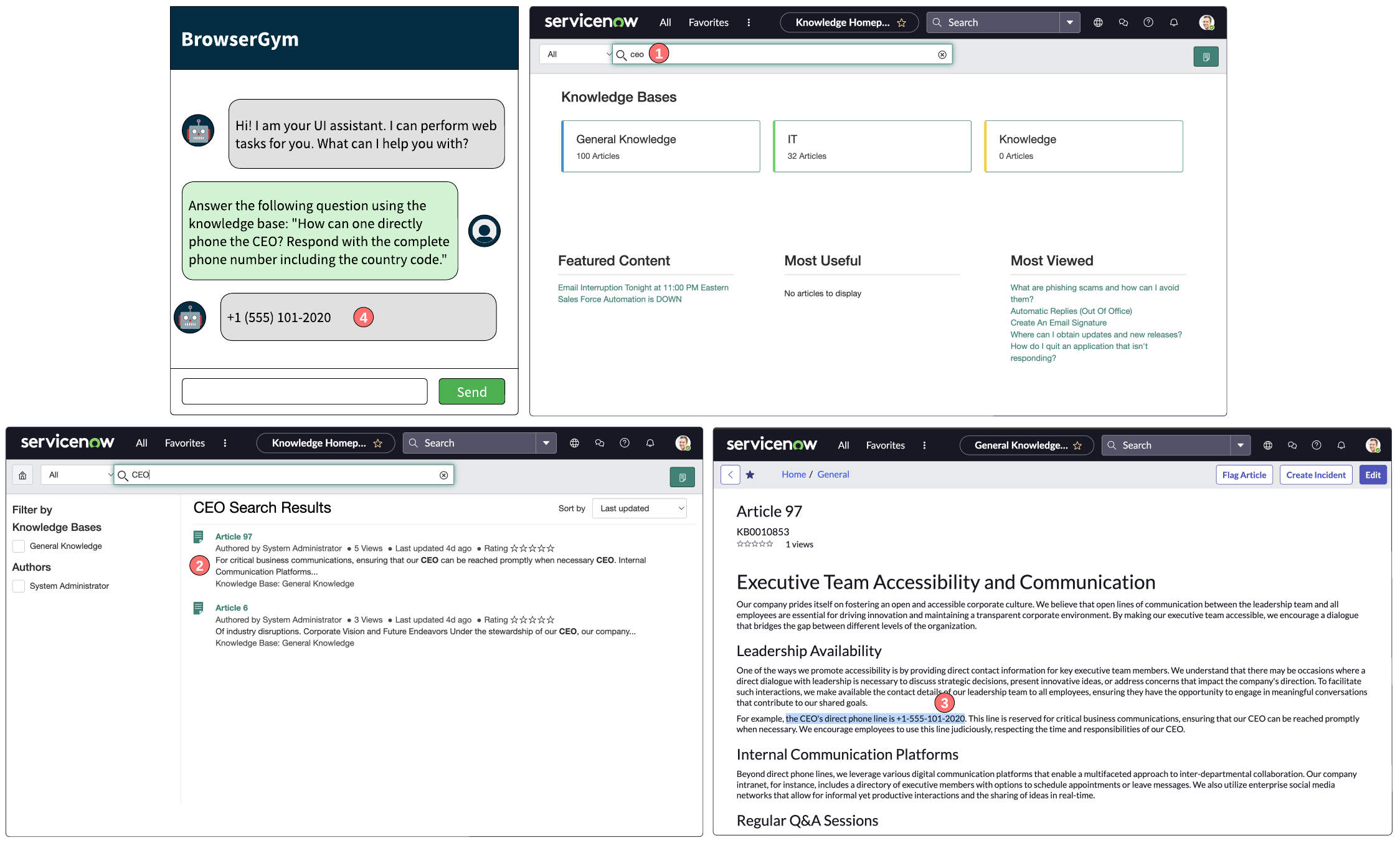}
    \caption{Example ``KnowledgeBaseSearch" Task -- The goal is given to the agent in natural language. As can be seen, the goal is designed to be very explicit, clearly stating which question must be answered and the expected format. Here, the agent must conduct a search using the search bar~\circledigit{1}. It must then browse all resulting articles~\circledigit{2} and read their content in order to find the desired information~\circledigit{3}. Finally, it must return this information to the user via the chat box for validation~\circledigit{4}.}
    \label{fig:wa-kbsearch-ui}
\end{figure*}

\begin{figure*}
    \centering
    \includegraphics[width=0.95\linewidth]{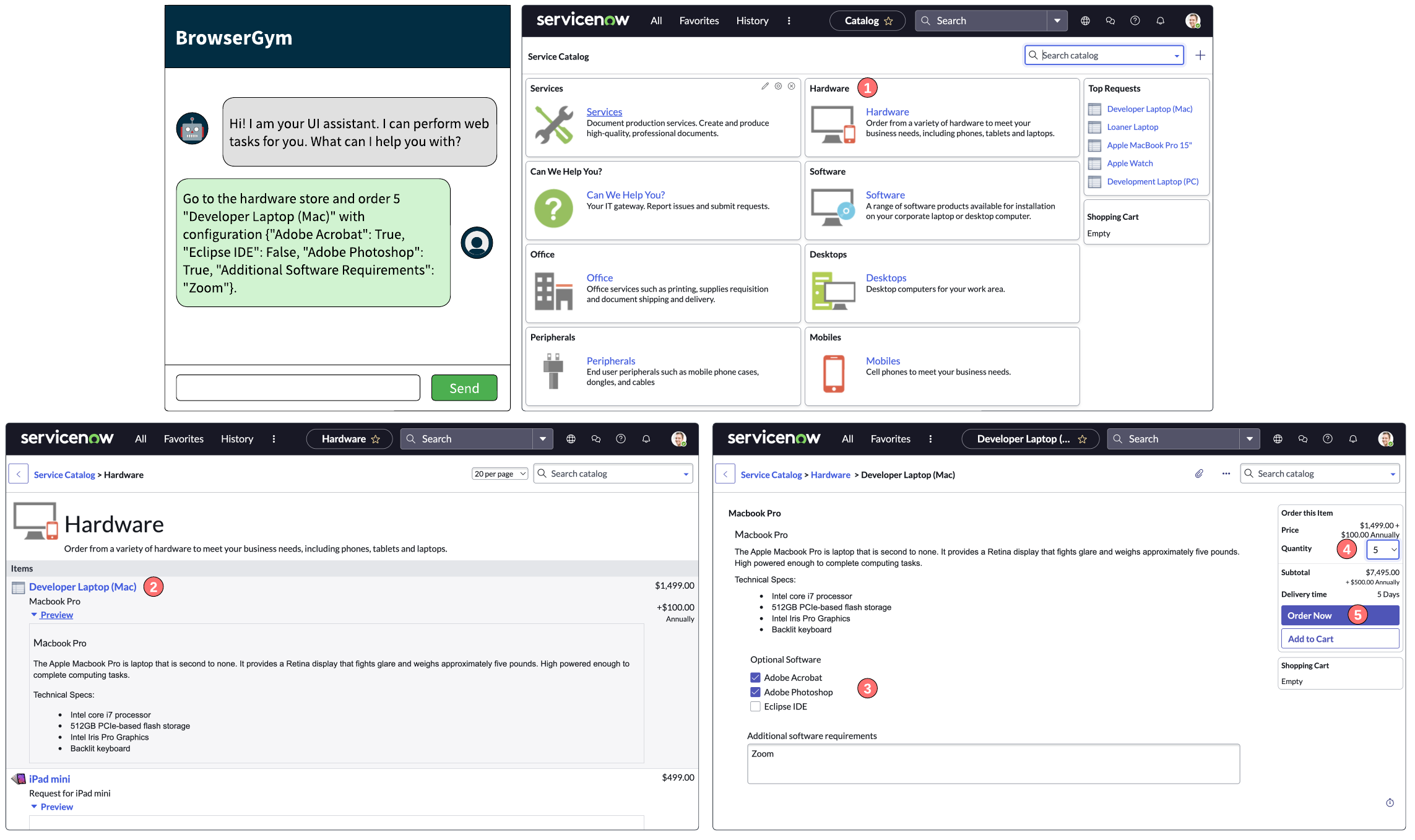}
    \caption{Example ``OrderDeveloperLaptopMac" Task -- The goal is given to the agent in natural language. As can be seen, the goal is designed to be very explicit, leaving no ambiguity on the task to perform. Here, the agent must navigate the service catalog to reach the appropriate item~\circledigit{1}--\circledigit{2}. Then, it must select the appropriate configuration~\circledigit{3} and quantity~\circledigit{4}. Finally, it must submit the order by clicking on the ``Order Now'' button.}
    \label{fig:wa-servicecatalog-ui}
\end{figure*}

\begin{figure*}
    \centering
    \includegraphics[width=0.95\linewidth]{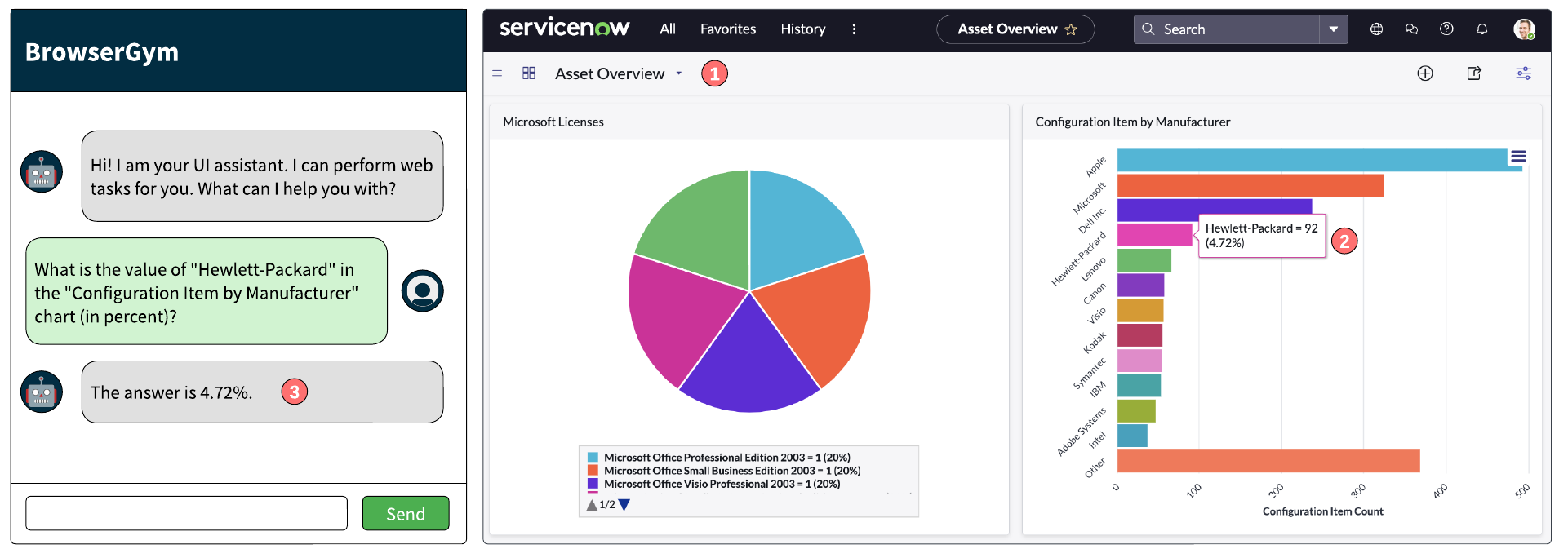}
    \caption{Example ``MultiChartValueRetrieval" Task -- The goal is given to the agent in natural language. As can be seen, the goal is designed to be very explicit, clearly stating which question must be answered and the expected format. Here, the agent must scan the entire dashboard to locate the relevant plot~\circledigit{1}. It must then search for the requested label and retrieve its value in the desired format (count or percentage)~\circledigit{2}. Finally, it must return this information to the user via the chat box for validation~\circledigit{3}.}
    \label{fig:wa-dashboard}
\end{figure*}

\begin{figure*}
    \centering
    \includegraphics[width=0.95\linewidth]{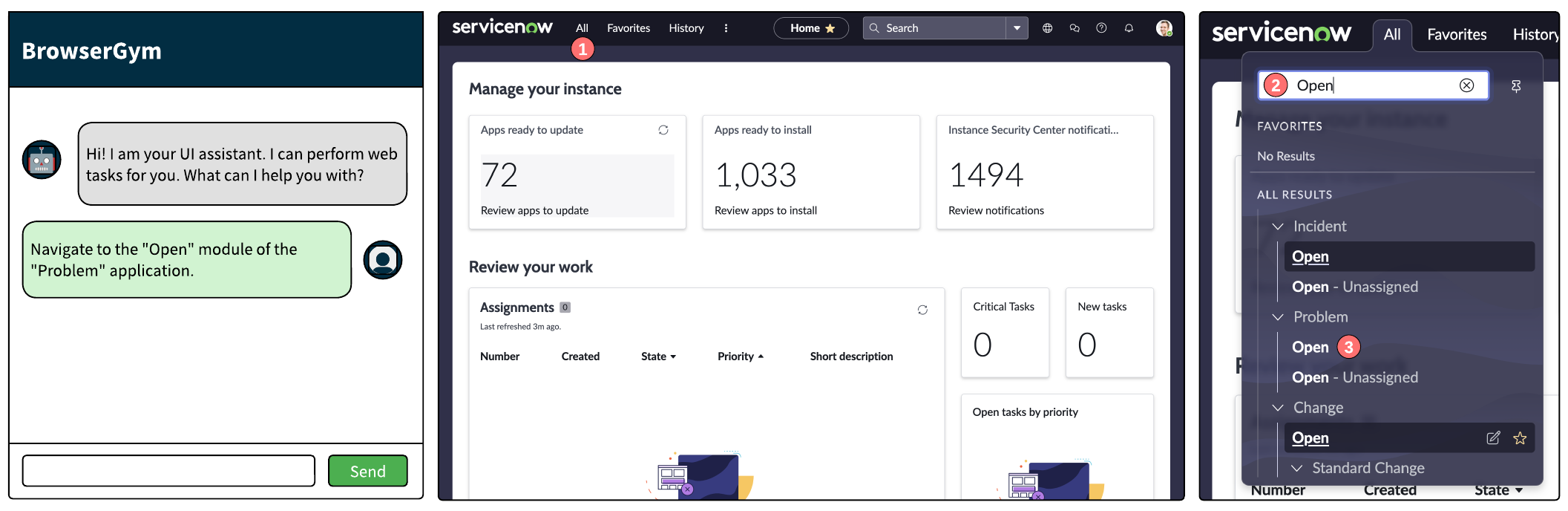}
    \caption{Example ``AllMenu" Task -- The goal is given to the agent in natural language. As can be seen, the goal is designed to be very explicit, leaving no ambiguity on the task to perform. Here, the agent must access the ``All'' menu~\circledigit{1}, conduct a search~\circledigit{2}, and select the right module~\circledigit{3}. As an alternative to~\circledigit{2}, the agent could scroll through the list. In this example, the agent must exercise caution when selecting the menu item to click, as many applications have an ``Open'' module.}
    \label{fig:wa-allmenu-ui}
\end{figure*}

In this section, we provide an example of the typical user interface encountered for each category of task (\cref{fig:wa-filterlist-ui,fig:wa-kbsearch-ui,fig:wa-servicecatalog-ui,fig:wa-allmenu-ui,fig:wa-dashboard}). We omit ``form'' tasks, as such an example has already been presented in \cref{fig:wa-form-trajectory}.

\FloatBarrier

\subsection{Knowledge Base Tasks -- Additional Details}\label{app:wa-kb-task-details}

This task consists of searching a company knowledge base for specific information to answer a given question.
Here, we explain how the knowledge base included in WorkArena is generated, how we produce the questions and answers used in each task instance, and how validation is performed.

\begin{figure}[!ht]
    \centering
    \setlength{\fboxsep}{0mm}
    \fbox{\includegraphics[width=0.99\textwidth, trim=0 0 37 0, clip]{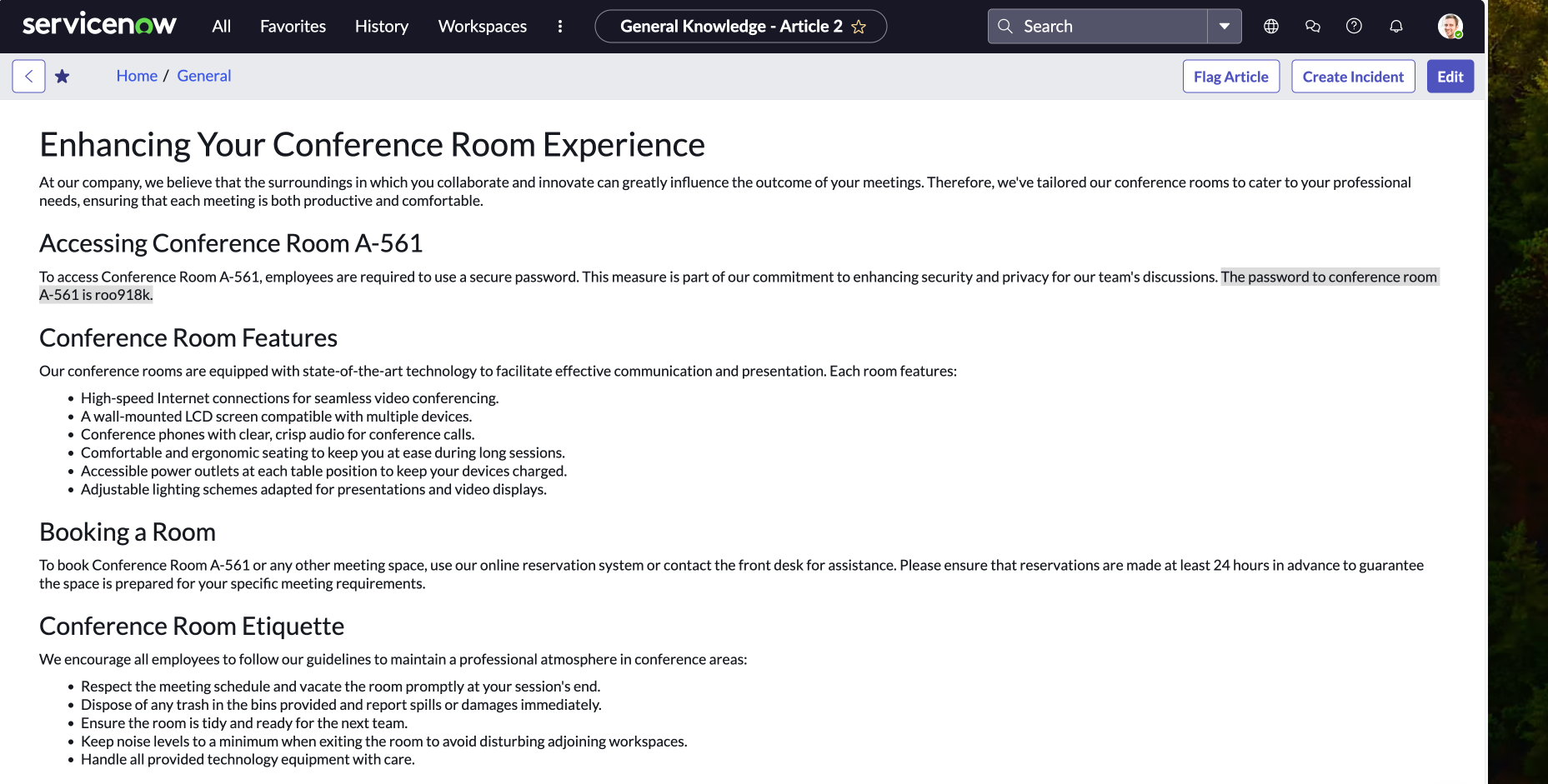}}
    \caption{Example of a generated knowledge base article. The fact (``password to conference room A-561'', ``roo918k'') is highlighted.}
    \label{fig:wa-kb-article}
\end{figure}

\paragraph{Generating the knowledge base:} The knowledge base included in WorkArena consists of 100 articles generated using GPT-4~\citep{OpenAI2023GPT4TR}. To achieve this, we start from a list of 100 facts, which are each composed of an \emph{item} and a \emph{value}. \cref{tab:wa-kb-facts} shows a few examples. Then, for each fact, we use GPT-4 to produce an article in HTML format and make sure that the exact string ``the \{fact\} is \{item\}'' is included in the article. An example article is shown in \cref{fig:wa-kb-article}.

\begin{table}[!ht]
    \centering
    \caption{Example facts included in the WorkArena knowledge base}
    \begin{tabular}{ll}
        \textbf{Fact}  &  \textbf{Item}\\ \toprule
       Password to conference room A-561  &  roo918k\\
       Address of office \#456  & 42, Pizza street, New York, USA\\
       CEO's name & Alex Johnson
    \end{tabular}
    \label{tab:wa-kb-facts}
\end{table}

\paragraph{Generating questions:} For each fact, i.e., (item, value) pair, we produce a list of questions that ask about \emph{item} and whose answers are exactly \emph{value}. We achieve this by prompting GPT-4 with the initial question ``What is \{item\}?'' and ask it to produce 10 alternative wordings for the question, as well as formatting instructions, to ensure that the answer is exactly \emph{value}. Then, we prompt GPT-3.5 with the generated article and each question, ensuring that every single one is answered correctly. If the model fails to answer a question, we ask GPT-4 to improve it and we repeat the process. Note that we use GPT-3.5 to answer the questions to avoid the pitfall where GPT-4 would cater to itself, producing ambiguous questions that it somehow succeeds in answering correctly. Example questions and formatting instructions are shown in \cref{tab:wa-kb-questions}.

\paragraph{Answer validation:} Despite the precise formatting instructions included with each question, we allow for slight variations in formatting and wording by verifying if the answer produced by the agent is within a set of acceptable answers. To produce such alternative answers, we provide the expected value to GPT-4 and ask it to produce 10 alternative formats. We include multiple examples in the prompt and inspect the results to ensure coherence. An example is shown in \cref{tab:wa-kb-answers}.

\begin{table}[!ht]
    \centering
    \caption{Example questions and formatting instructions produced for initial question ``What is the address of Office \#456?''}
    \begin{adjustbox}{max width=\textwidth}
    \begin{tabular}{ll}
          \textbf{Question} & \textbf{Formatting Instructions}\\ \toprule
          Could you provide the street location for Office \#456? & Make sure to include the Street Number, Street Name, City, and Country.\\
          Where can Office \#456 be found? & Provide the exact street address, city, and country.\\
          Where should one go to visit Office \#456? & Please respond with the format: Number, Street, City, Country.\\
          What's the precise location of Office \#456? & Answer with the Street Number, Street Name, City, and Country.
    \end{tabular}
    \end{adjustbox}
    \label{tab:wa-kb-questions}
\end{table}

\begin{table}[!ht]
    \centering
    \caption{Example alternative answers for question ``What is the address of Office \#456?'', where the expected answer is ``42, Pizza street, New York, USA''.}
    \begin{tabular}{l}
          \textbf{Alternative Answer}\\ \toprule
          42 Pizza Street, New York, USA\\
          42, Pizza St., NY, United States\\
          \#42 Pizza Street, New York, U.S.\\
          42 Pizza St, New York City, United States of America
    \end{tabular}
    \label{tab:wa-kb-answers}
\end{table}

\FloatBarrier

\clearpage

\section{BrowserGym -- Additional Details}\label{app:browsergym}

\subsection{Action Space}
\begin{table}[ht]
    \centering
    \footnotesize
    \caption{The complete action space of BrowserGym.}
    \label{tab:browsergym_action_space}
    \begin{tabular*}{\textwidth}{@{\extracolsep{\fill}}MMp{2.75in}@{}}
        \toprule
        {\bf Category} & {\bf Primitive} & {\bf Description} \\
        \toprule
        \multicolumn{1}{M}{\multirow{7}{*}{bid}} & fill(bid, text) & Fill an input field with text.\\
        & click(bid, button) & Click an element. \\
        & dblclick(bid, button) & Double-click an element. \\
        & hover(bid) & Hover the mouse over an element. \\
        & press(bid, key\_comb) & Focus an element and press a combination of keys. \\
        & focus(bid) & Focus an element. \\
        & clear(bid) & Clear an input field. \\
        & select\_option(bid, options) & Select one or multiple options in a drop-down element. \\
        & drag\_and\_drop(from\_bid, to\_bid) & Drag and drop one element to another. \\
        \midrule
        \multicolumn{1}{M}{\multirow{10}{*}{coord}} & mouse\_move(x, y) & Move the mouse to a location. \\
        & mouse\_down(x, y, button) & Move the mouse to a location then press and hold a mouse button. \\
        & mouse\_up(x, y, button) & Move the mouse to a location then release a mouse button. \\
        & mouse\_click(x, y, button) & Move the mouse to a location and click a mouse button. \\
        & mouse\_dblclick(x, y, button) & Move the mouse to a location and double-click a mouse button. \\
        & mouse\_drag\_and\_drop(from\_x, from\_y, to\_x, to\_y) & Drag and drop from a location to a location. \\
        & keyboard\_down(key) & Press and holds a keyboard key. \\
        & keyboard\_up(key) & Release a keyboard key. \\
        & keyboard\_press(key\_comb) & Press a combination of keys. \\
        & keyboard\_type(text) & Types a string of text through the keyboard. \\
        & keyboard\_insert\_text(text) & Insert a string of text in the currently focused element. \\
        \midrule
        \multicolumn{1}{M}{\multirow{3}{*}{tab}} & new\_tab() & Open a new tab. \\
        & tab\_close() & Close the current tab. \\
        & tab\_focus(index) & Bring a tab to front (activate tab). \\
        \midrule
        \multicolumn{1}{M}{\multirow{3}{*}{nav}} & go\_back() & Navigate to the previous page in history. \\
        & go\_forward() & Navigate to the next page in history. \\
        & goto(url) & Navigate to a url. \\
        \midrule
        \multicolumn{1}{M}{\multirow{2}{*}{misc}} & scroll(dx, dy) & Scroll pixels in X and/or Y direction. \\
        & send\_msg\_to\_user(text) & Send a message to the user in the chat. \\
        & noop() & Do nothing. \\
        \midrule
        \multicolumn{1}{M}{\multirow{1}{*}{python}} & \textrm{Any python code (\textsc{unsafe!})} & Executes code with playwright, the active \texttt{\footnotesize page} and the \texttt{\footnotesize send\_msg\_to\_user(text)} primitive available. \\
        \bottomrule
    \end{tabular*}
\end{table}

\subsection{MiniWoB}\label{app:miniwob}

As part of BrowserGym, we provide a port of the MiniWoB benchmark~\citep{Shi2017MiniWoB, LiuL18MiniWoB}. Whereas in MiniWoB the goal of each task is embedded in the HTML of the web page, in BrowserGym the goal is provided in a separate chat window, accessible to the agent. We therefore run a minimal Javascript snippet in the \texttt{setup()} function of MiniWoB tasks to extract the goal from the web page and place it in the chat instead. Apart from that change and removing the hard time limit per episode, porting MiniWoB to BrowserGym required minimal effort.

\begin{figure}
    \centering
    \includegraphics[width=0.99\textwidth]{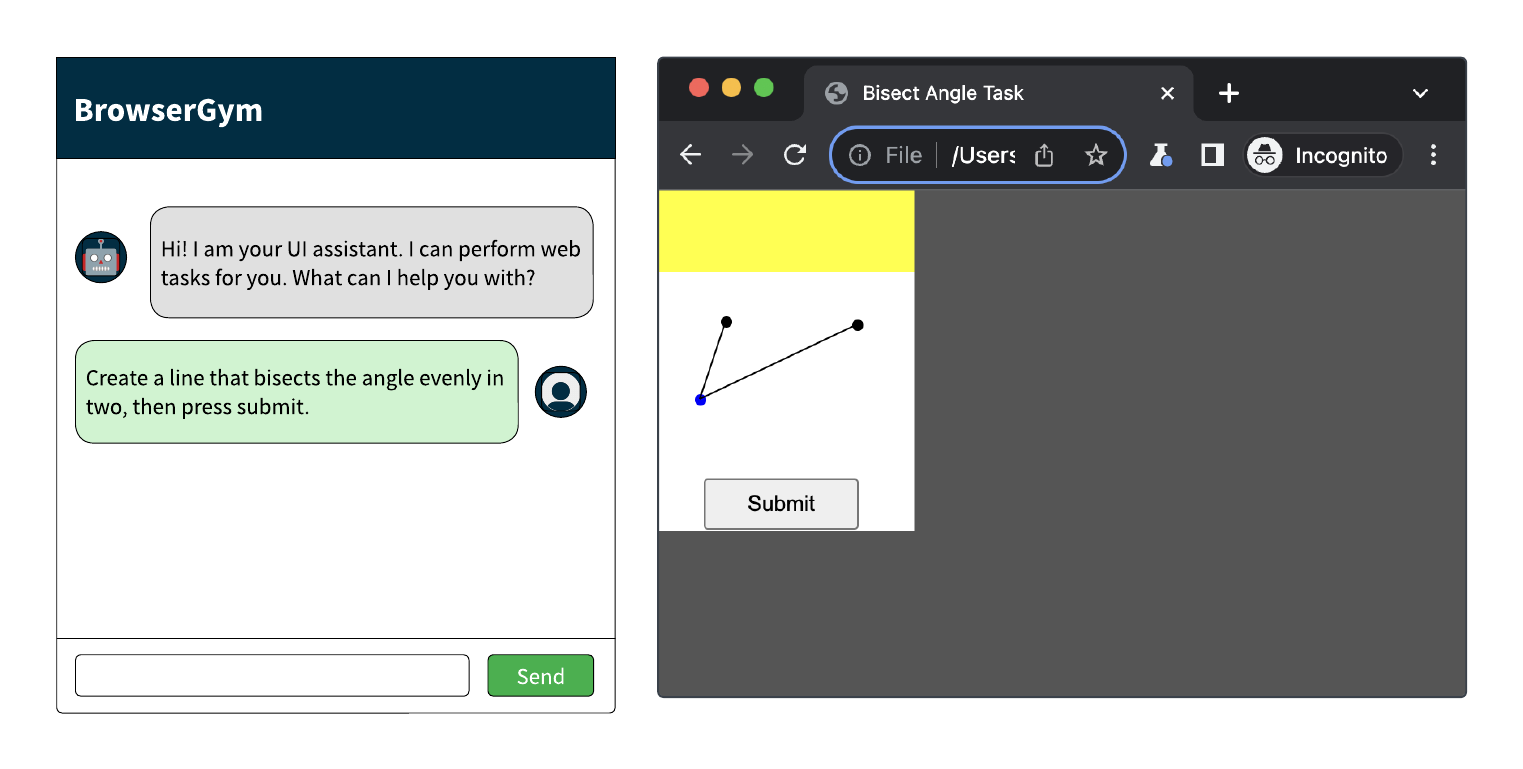}
    \caption{Example of a MiniWoB task as rendered in BrowserGym.}
    \label{fig:browsergym_miniwob}
\end{figure}

\clearpage

\end{document}